\documentclass{article}

\usepackage{microtype}
\usepackage{graphicx}
\usepackage{caption}
\usepackage{subcaption}
\usepackage{booktabs} %
\usepackage{makecell}
\usepackage{adjustbox}
\usepackage{multirow}
\usepackage{algorithm}
\usepackage{algorithmic}
\usepackage{xspace}
\usepackage{tabularx}
\usepackage{xcolor}

\usepackage{hyperref}
\usepackage{amsmath,amsfonts,bm}

\usepackage[accepted]{icml2024}

\usepackage{amsmath}
\usepackage{amssymb}
\usepackage{mathtools}
\usepackage{amsthm}

\usepackage{siunitx} %

\usepackage[capitalize,noabbrev]{cleveref}

\theoremstyle{plain}

\theoremstyle{definition}

\theoremstyle{remark}

\newcommand*{\acc}[1]{\num[round-mode=places,round-precision=1]{#1}\%}

\newcommand{\punct}{Punct\xspace}
\newcommand{\puncttable}{Punct\xspace}
\newcommand{\gpt}{GPT-4\xspace}
\newcommand{\codellama}{Code~Llama\xspace}
\newcommand{\llama}{Llama\xspace}
\newcommand{\smallmodel}{NL\xspace}
\newcommand{\smallcodemodel}{Code\xspace}

\newcommand{\identity}{Identity\xspace}
\newcommand{\identitytable}{Identity\xspace}

\newcommand{\merged}{Merged\xspace}
\newcommand{\mergedtable}{Merged\xspace}

\font\Fulwidth=cmss12
\letterspacefont\fw\Fulwidth 500

\usepackage[textsize=tiny]{todonotes}

\DeclareMathOperator{\length}{length}

\begin{document}

\icmltitlerunning{Getting the most out of your tokenizer for pre-training and domain adaptation}

\twocolumn[

\icmltitle{Getting the most out of your tokenizer for pre-training and domain adaptation}

\begin{icmlauthorlist}
\icmlauthor{Gautier Dagan}{ed}
\icmlauthor{Gabriel Synnaeve}{meta}
\icmlauthor{Baptiste Rozière}{meta}
\end{icmlauthorlist}

\icmlaffiliation{ed}{University of Edinburgh, Edinburgh, UK}
\icmlaffiliation{meta}{Meta AI, Paris, France}
\icmlcorrespondingauthor{Gautier Dagan}{gautier.dagan@ed.ac.uk}

\icmlkeywords{Machine Learning, Deep Learning, Tokenization, Code Generation}

\vskip 0.3in
]

\printAffiliations{}

\newcommand{\changed}[1]{{\color{red}#1}}

\begin{abstract}
Tokenization is an understudied and often neglected component of modern LLMs. 
Most published works use a single tokenizer for all experiments, often borrowed from another model, without performing ablations or analysis to optimize tokenization. 
Moreover, the tokenizer is generally kept unchanged when fine-tuning a base model. 
In this paper, we show that the size, pre-tokenization regular expression, and training data of a tokenizer can significantly impact the model's generation speed, effective context size, memory usage, and downstream performance. 
We train specialized Byte-Pair Encoding code tokenizers, and conduct extensive ablations on the impact of tokenizer design on the performance of LLMs for code generation tasks such as HumanEval and MBPP, and provide recommendations for tokenizer hyper-parameters selection and switching the tokenizer in a pre-trained LLM. 
We perform our experiments on models trained from scratch and from pre-trained models, verifying their applicability to a wide range of use-cases.
We find that when fine-tuning on more than 50 billion tokens, we can specialize the tokenizer of a pre-trained LLM to obtain large gains in generation speed and effective context size.
\end{abstract}

\section{Introduction}

Tokenizers transform raw sequences of text into tokens, and are an essential part of most modern language models. 
They are generally built using the Byte-Pair Encoding (BPE) algorithm \citep{sennrich-etal-2016-neural}, and more rarely with the Unigram algorithm \cite{Kudo_2018}.
Building a tokenizer is one of the first steps of any language modeling project.
Modifying the tokenizer impacts everything downstream, and is generally more cumbersome than performing ablations on other hyper-parameters or modeling features. 
Hence, practitioners often report results based on a single set of tokenizer hyper-parameters.

\begin{figure}
    \centering
    \scalebox{0.9}{
    \includegraphics[width=\linewidth]{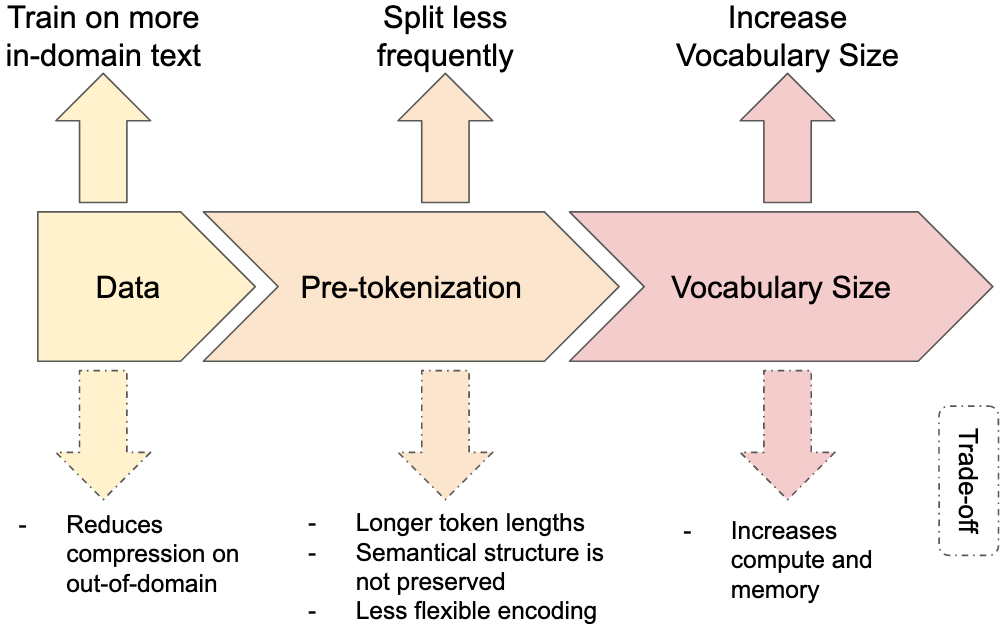}
    }
    \caption{Three ways to increase in-domain compression in a BPE tokenizer with their respective trade-offs.}
    \label{fig:compression_tradeoff}
\end{figure}

This is especially true for projects fine-tuning a pre-trained LLM for a specific task or domain.
The tokenizer of the base model is generally unchanged, leading to tokenizers being applied to domains for which they are sub-optimal, hurting efficiency and potentially the performance of the final model.
For instance, \codellama~\cite{roziere2023code} re-uses the tokenizer that was used by its base model, \llama~2~\cite{touvron2023llama2}, which uses the tokenizer from the original \llama model~\cite{touvron2023llama}. 
This means that \codellama, while being a popular model fine-tuned on a domain specific task, is still limited by the decisions taken during the pre-training of the original base model.
However, higher-compression code tokenizers exist.
For instance, the InCoder~\cite{incoder} tokenizer is trained on source code, has a larger vocabulary size, and uses less restrictive pre-processing rules.
As a result, it uses $25\%$ less tokens than the \llama tokenizer on average when encoding source code (see Table~\ref{tab:NSL}).

Efficiency in token usage can have a significant impact on LLM inference and training. 
In the context of modern LLMs, where the inference budget is non-negligible~\citep{touvron2023llama}, enhancements in compression like these lead to more efficient inference in both FLOPS and memory usage.
Compression also increases the effective context length of the model, defined as the number of characters the model can ingest on average.
Conversely, high-compression tokenizers can negatively impact the performance of the model in other ways.
While the compute cost of increasing the size of the vocabulary is negligible in practice for LLMs, it can significantly impact memory usage, especially for smaller models.
Compute is also critical limiting factor when training LLMs, and high-compression tokenizers allow to train on more text for a fixed compute budget.

Existing tokenizer studies perform experiments at a much smaller scale than what is typical for modern LLMs~\cite{gowda2020finding,Chirkova_Troshin_2022}, or focus on multilingual tasks \cite{rust-etal-2021-good,limisiewicz-etal-2023-tokenization,Zouhar_Meister_Gastaldi_Du_Sachan_Cotterell_2023}.
It is not clear whether these findings transfer to models with greater capacity and trained on more tokens.
Similarly, some previous work has looked at adapting tokenizers in pre-trained LLMs \cite{vipi,FVT,gee-etal-2023-multi}, but only on smaller LLMs ($<1$B parameters) and vocabulary sizes ($<32$k tokens).

In this paper, we focus our experiments on modern code LLMs of 1.5B and 7B parameters. %
Our contributions are as follows: 
\begin{itemize}
    \item We compare popular code tokenizers, clarifying their respective performances and trade-offs made.
    \item We study the impact of vocabulary size, pre-tokenization regular expression on compression and downstream code generation performance when fine-tuning and training from scratch. We observe that the pre-tokenization can substantially impact both metrics and that vocabulary size has little impact on coding performance.
    \item For fine-tuning existing models, we show that the tokenizer can be changed with little impact to downstream performance when training on 50B tokens or more. %
\end{itemize}

In section~\ref{sec:compression-tradeoffs}, we detail three ways to increase tokenizer compression, and propose methods to calculate inference and memory optimal vocabulary sizes.
In section~\ref{sec:experiments}, we study the impact of fine-tuning and training from scratch an LLM with a different tokenizer, and evaluate the effects of tokenizer settings on downstream code generation. 

\begin{table}[h]
\scalebox{0.75}{
\begin{tabular}{lrrrrr}
\toprule
\multicolumn{2}{r}{} & \multicolumn{4}{c}{NSL (↓)} \\
 & \makecell{Size} & Avg. & Code & Eng. & Mult. \\
\midrule
\small{GPT-2 \cite{gpt2}} & 50k & 1.13 & 1.19 & 0.86 & 1.33 \\
\small{DeepSeek Coder \cite{deepseekai_2023_deepseekcoder13binstruct}} & 32k & 1.06 & 1.00 & 0.98 & 1.19 \\
\small{\llama \cite{touvron2023llama}} & 32k & 1.00 & 1.00 & 1.00 & 1.00 \\
\small{CodeGen \cite{Nijkamp_Pang_Hayashi_Tu_Wang_Zhou_Savarese_Xiong_2023}} & 50k & 1.05 & 0.95 & 0.86 & 1.33 \\
\small{CodeT5 \cite{Wang_Wang_Joty_Hoi_2021}} & 32k & 1.29 & 0.94 & 1.11 & 1.83 \\
\small{SantaCoder \cite{santacoder}} & 49k & 1.04 & 0.88 & 1.07 & 1.17 \\
\small{StarCoder \cite{starcoder}} & 49k & 0.99 & 0.87 & 1.04 & 1.07 \\
\small{Replit Code \cite{replit}} & 32k & 1.00 & 0.85 & 1.06 & 1.10 \\
\small{GPT-4 \cite{openai2023gpt4}} & 100k & 0.85 & 0.75 & 0.84 & 0.95 \\
\small{InCoder \cite{incoder}} & 50k & 1.03 & 0.74 & 1.02 & 1.31 \\
\hline
\multicolumn{6}{l}{Ours}\\
\small{\puncttable} & 32k & 0.98 & 0.86 & 0.96 & 1.11 \\
\small{\puncttable} & 64k & 0.90 & 0.82 & 0.89 & 0.99 \\
\small{\puncttable} & 80k & 0.88 & 0.81 & 0.88 & 0.95 \\
\small{\puncttable} & 100k & 0.86 & 0.81 & 0.86 & 0.92 \\
\small{\gpt} & 32k & 0.97 & 0.81 & 0.97 & 1.13 \\
\small{\gpt} & 64k & 0.89 & 0.76 & 0.90 & 1.01 \\
\small{\gpt} & 80k & 0.87 & 0.75 & 0.88 & 0.98 \\
\small{\gpt} & 100k & 0.85 & 0.74 & 0.86 & 0.94 \\
\small{\identitytable} & 32k & 0.92 & 0.69 & 0.89 & 1.16 \\
\small{\identitytable} & 64k & 0.82 & 0.63 & 0.79 & 1.04 \\
\small{\identitytable} & 80k & 0.80 & 0.61 & 0.76 & 1.01 \\
\small{\identitytable} & 100k & 0.77 & 0.59 & 0.74 & 0.98 \\
\small{\mergedtable} & 80k & 0.90 & 0.80 & 0.95 & 0.94 \\
\bottomrule
\end{tabular}
}
\caption{Comparison of popular code tokenizers by their Normalized Sequence Length (NSL), in this case NSL calculated against the \llama tokenizer, and indicates the average tokenized sequence length that a tokenizer would produce compared to \llama (see section~
\ref{sec:metrics}). The lower the NSL, the more efficient the tokenizer is at compressing the dataset. For example, on our Code subset, the InCoder \cite{incoder} tokenizer uses on average $26\%$ less tokens than the \llama tokenizer.
}
\label{tab:NSL}

\end{table}

\section{Compression trade-offs}
\label{sec:compression-tradeoffs}

We identify three main levers impacting the downstream compression of a tokenizer on a specific domain (see Figure~\ref{fig:compression_tradeoff}). 
The first one is the data used to train the tokenizer, where using data sampled from the in-domain distribution will increase in-domain compression. 
The second lever is the pre-tokenization scheme, which can be written as a regular expression defining how the text is split before it is passed to the BPE tokenizer. %
Splitting sequences prevents BPE from merging certain tokens, for instance splitting on white spaces means that a token cannot span two space-separated words.
It leads to shorter tokens and thus worse compression rates, but is generally done to improve downstream performance. %
Finally, increasing the vocabulary size leads to higher compression at the cost of compute and memory.

It is important to note that higher compression rates could also lead to deteriorated downstream performance, since shorter sequences give less effective FLOPs to a model to reason \cite{pause_tokens}.
This is a consequence of the modern Transformer decoder architecture in which \textit{every token requires an additional forward pass to generate}.
Therefore even seemingly low-information tokens might still provide gains on downstream task.
This is evidenced by \citet{pause_tokens}, who propose Pause Tokens, special empty tokens added to the context to enable the model to `pause' its reasoning and add FLOPs during inference.

\subsection{Compression metrics}
\label{sec:metrics}

Compression is always measured as a ratio of a quantity with respect to another.
We measure two compression metrics, the first, Normalized Sequence Length (NSL), compares the compression of a given tokenizer with respect to our baseline \llama tokenizer. %
The second, is the average number of Bytes per Token, and is calculated by dividing the number of UTF-8 bytes by the number of tokens produced by the tokenizer on a given text.

Formally, we define NSL $c_{\frac{\lambda}{\beta}}$ as the ratio between the length of an encoded sequence from a tokenizer $T_\lambda$ and a tokenizer $T_\beta$. For $N$ examples taken from a dataset $D$:

$$
    c_{\frac{\lambda}{\beta}} = \frac{\sum_{i=1}^N \length(T_\lambda(D_i))}{\sum_{i=1}^N \length(T_\beta(D_i))}
$$

We use the \llama tokenizer \cite{touvron2023llama} as our reference $T_\beta$.
In other words, if $T_\lambda$ has an NSL of $0.75$ on average, it means that sequences encoded with $T_\lambda$ contain 25\% less tokens on average than those encoded with \llama.

We use the public datasets CCnet \cite{ccnet}, Wikipedia, and the Stack \cite{Kocetkov2022TheS3} for tokenizer training and evaluation.
We divide data into three categories: English, code, and multilingual data (see Appendix~\ref{sec:data}).
Multilingual data consists of non-English texts from Wikipedia and includes 28 languages.
The code category includes 30 different programming languages.
For each subset (programming language, natural language) of each dataset, we hold-out 1000 examples (files) and use these to compute the compression ratio.
The compression ratio of a category is calculated as an average over all its respective subsets.
This ensures that subsets which contains longer sequences, for instance C++ code, is weighted equally against subsets that contain shorter sequences in average.
The overall average NSL is calculated as the average over the held-out code, English and multilingual data.

\begin{figure}
    \centering
    \includegraphics[width=\linewidth]{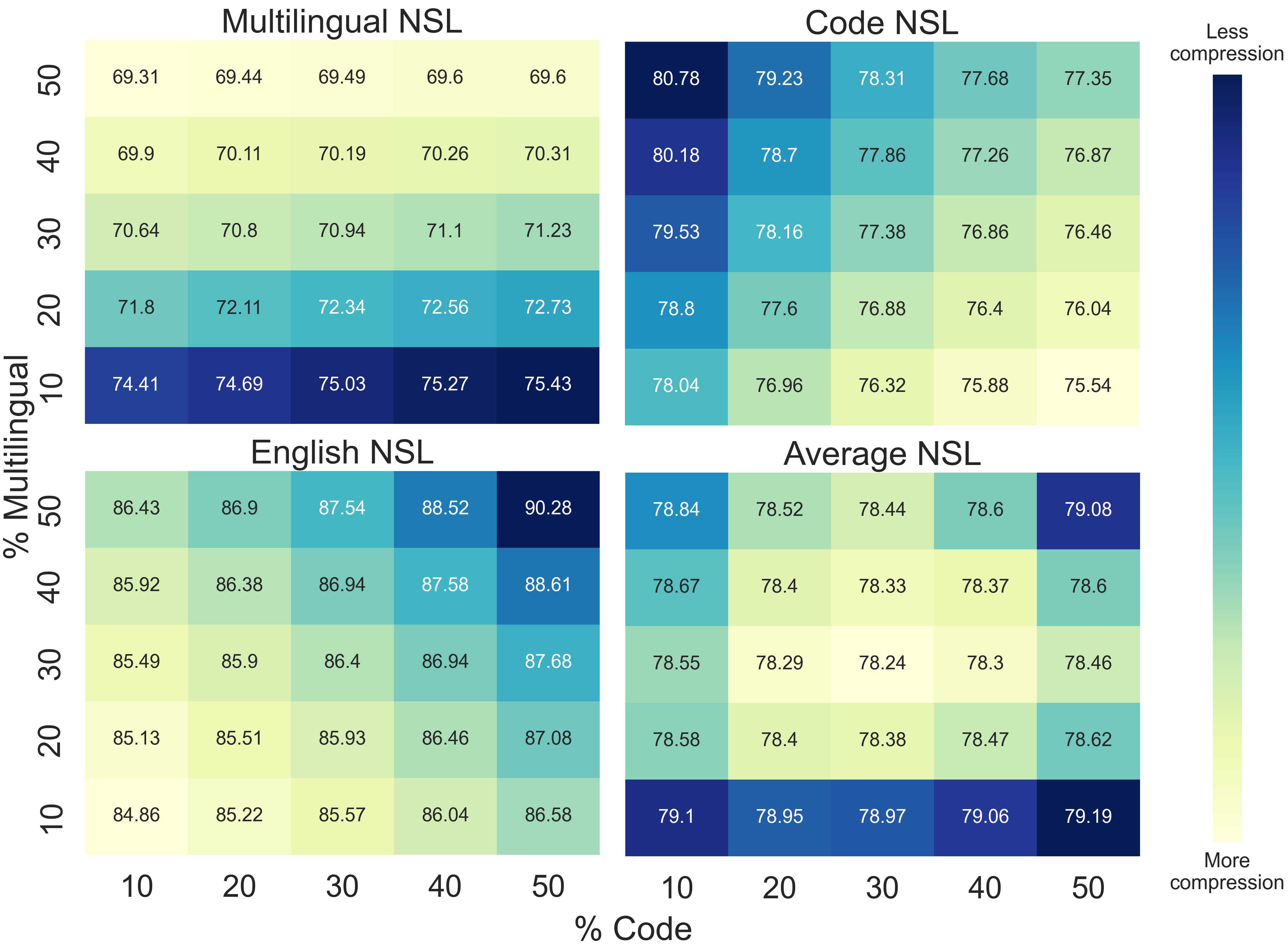}
    \caption{Tokenizers trained with different \% of code, English, multilingual data. Unsurprisingly, training on code improves code compression, training on multilingual data improves multilingual compression, and training on an even mix of all three subset leads to the best average compression.}
    \label{fig:compression_heatmap}
\end{figure}

\subsection{Algorithm} %

We use the BPE \cite{sennrich-etal-2016-neural} tokenization algorithm as it is the most commonly used to train general and code-specific tokenizers.
We considered two popular libraries implementing BPE training, Google's Sentencepiece~\citep{Kudo2018SentencePieceAS} and HuggingFace tokenizers~\citep{Wolf2019HuggingFacesTS}.
Since we measure the effects of different pre-tokenization schemes on code, we opt to use HuggingFace's tokenizer library as it supports a regular expression-based pre-tokenization and better handles special formatting characters such as tabs and new lines.

\subsection{Data}
Perhaps unsurprisingly, the data used to train a BPE tokenizer significantly impact its compression on evaluation datasets.
We train tokenizers on different dataset mixes and compare the compression (NSL) obtained on a held-out sets. 
We fix the number of characters used to train learn the BPE tokenizer to 10~Billion, and vary only the percentage of code and multilingual training data in the training dataset. 
We keep all the other hyper-parameters constant. %

Figure~\ref{fig:compression_heatmap} shows the NSL of our trained tokenizers on three held-out sets for Multilingual, Code and English.
We measure the average NSL on all subset, as well as the average over the three (Average NSL).
As expected, we find that training on more code improves code compression, training on multilingual data improves multilingual compression, and training on an even mix of all three subset leads to the best global average compression.
Figure~\ref{fig:compression_heatmap} reinforces the notion that \textit{tokenizers should be trained on the data mix that they are expected to see during training/inference}. 
We also observe that the NSL on any given subset only varies by 5 to 6 percentage points. 
For example when 50\% of the data is multilingual, the NSL is only improved by about $5$\% compared to when 10\% of the data is multilingual.

For the rest of this paper, since our target domain is Code Generation, we train all tokenizers (shown in Table~\ref{tab:NSL}) on a data distribution of 70\% code and 30\% English.

\begin{figure}
    \centering
    \includegraphics[width=\linewidth]{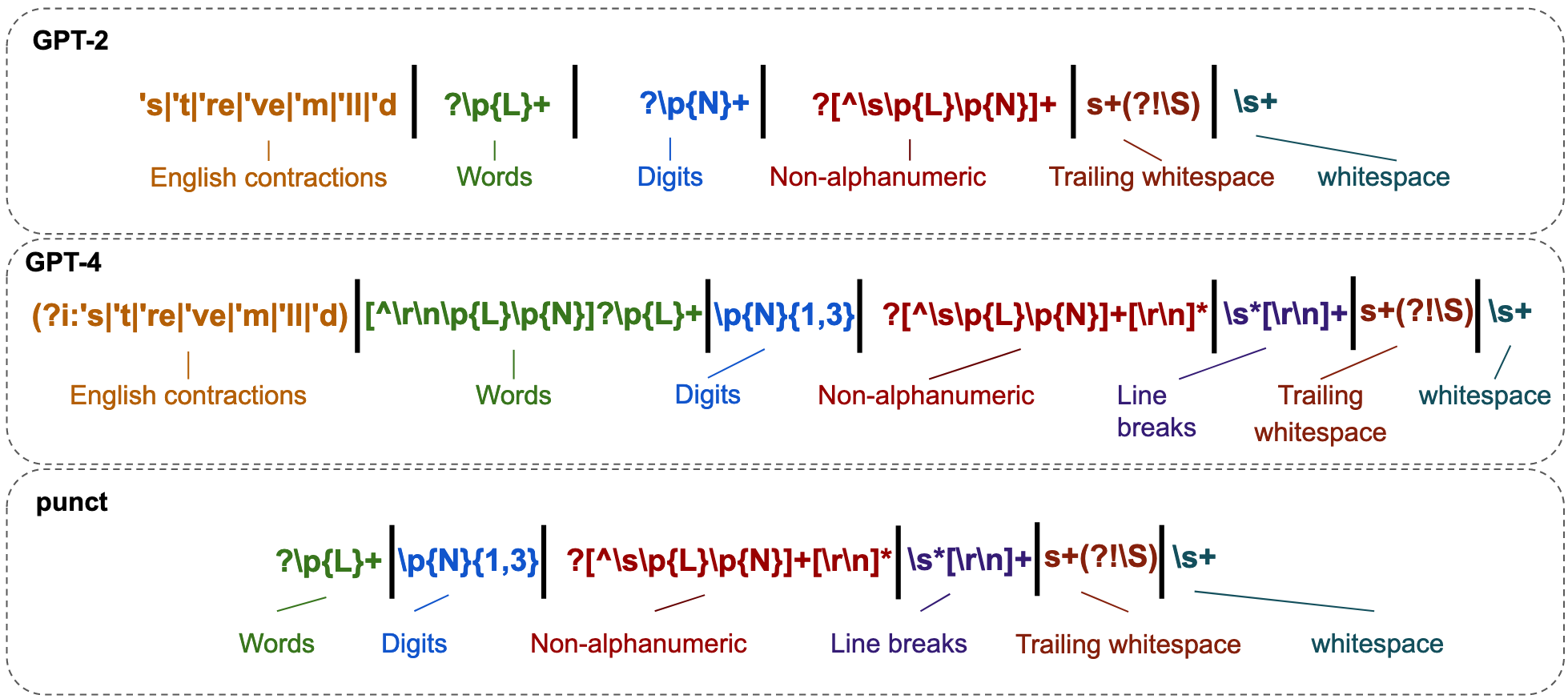}
    \caption{The GPT-2 \cite{gpt2} and GPT-4 \cite{openai2023gpt4} pre-tokenization regular expressions decomposed into functional sub-parts, and another version dubbed \punct which we introduce to ablate some of the changes introduced in GPT-4.
    \punct does away with the English-specific contractions and prevents certain whitespace and punctuation tokens such as \texttt{\textbackslash t} or \texttt{.} to be encoded at the start of an alpha-only token (see Appendix~\ref{sec:example} for an example).
    }
    \label{fig:regexex}
\end{figure}

\begin{figure*}
    \centering
    \includegraphics[width=\linewidth]{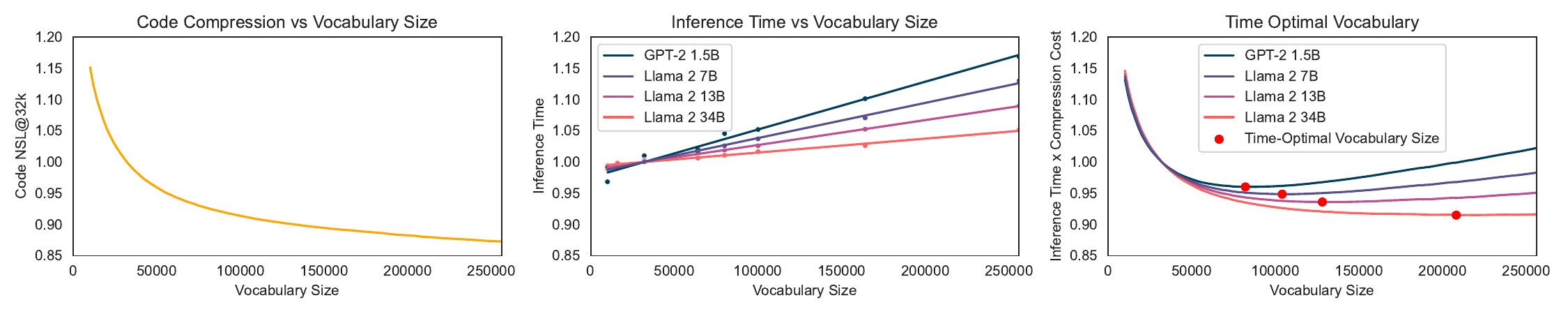}
    \includegraphics[width=\linewidth]{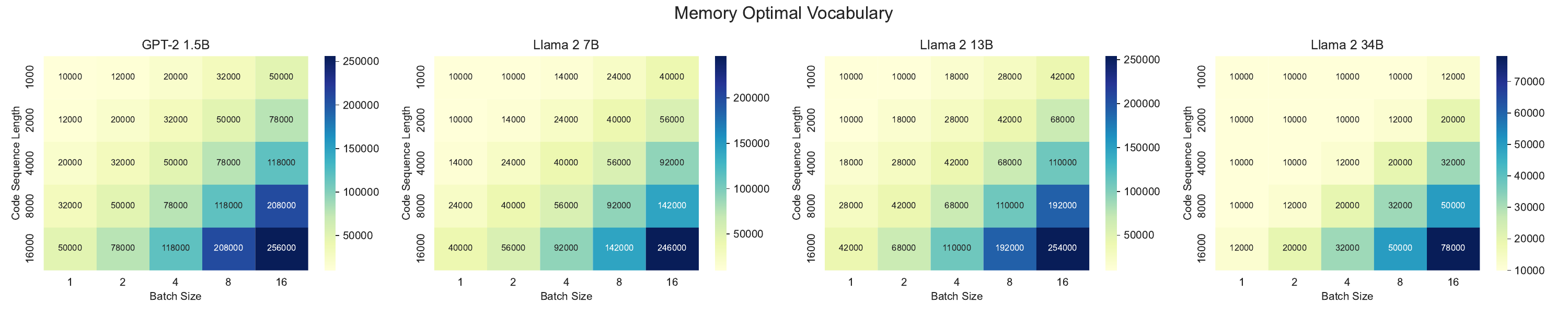}
    \caption{
    \textbf{(top left)} For given fixed set of tokenizer settings, we measure the Code NSL of different vocabulary sizes. We set the reference point to the tokenizer trained @32k tokens to compare against.
    \textbf{(top middle)} We measure the inference time for a set of vocabulary sizes and models with a fixed sequence length of 4096, and plot a linear regression over observations. We normalize predictions to a vocab of 32k. 
    \textbf{(top right)} By combining the compression and inference time trade-offs, we obtain a simple cost function that describes an optimal inference time.
    \textbf{(bottom)} We use equation~\ref{eq:eq1} to find the memory optimal vocabulary size for different models. \llama~2~34B uses grouped-query attention, which significantly reduces the cache's memory usage and the memory-optimal vocabulary size.
    }
    \label{fig:parameter-optimal}
\end{figure*}

\subsection{Pre-tokenization}

BPE \cite{sennrich-etal-2016-neural} operates by iteratively merging frequent adjacent characters or character sequences to build a vocabulary from sub-word units. 
When applied naively, this method leads to tokens forming around common phrases or sentences, which could be sub-optimal for certain tasks.
\textit{Pre-tokenization} is a pre-processing step that happens before passing the text to the tokenization algorithm.
Most commonly, this step involves breaking down text into more granular chunks.
This can be based on linguistic rules, such as splitting on punctuation marks or spaces, to ensure that the individual learned tokens are meaningful and promote compositional re-use.

For instance, consider that without any pre-tokenization, entire phrases or common occurrences such as dates (\texttt{2022}) can be represented as a single token.
While this might be optimal in terms of compression, this introduces complexities for the downstream LLM.
For instance, if tasked with an arithmetic problem, arbitrary tokens such as \texttt{2022} forces the model to learn arithmetic for every token independently.
In comparison, a tokenizer splitting on individual digits would encode \texttt{2022} as \texttt{2}, \texttt{0}, \texttt{2}, \texttt{2} separately, and could therefore learn to generalize basic arithmetic.
Previous works have also shown that digit tokenization can significantly impact arithmetic performance~\citep{nogueira2021investigating,thawani-etal-2021-representing}.

Note that to keep our tokenization scheme perfectly reversible, we abstain from any form of normalization in our pre-tokenization step (see Appendix~\ref{sec:normalization}). 

\paragraph{Pre-tokenizers based on regular expressions.}
Regular expressions provide a powerful mechanism for defining patterns for text segmentation.
They are often used to create custom pre-tokenizers tailored to specific data or tasks.
GPT-2 \cite{gpt2}, in particular, introduced a large regular expression (shown in Figure~\ref{fig:regexex}) to split the text into chunks before applying BPE.
In GPT-4 \cite{openai2023gpt4}, that regular expression was expanded to capture more specific groups, for instance limiting the number of digits allowed in a token to three instead of an unbounded number.
Figure~\ref{fig:regexex}) presents a detailed breakdown of the regular expression used in GPT-2, GPT-4, and a simplified regular expression we refer to as \punct (see Appendix~\ref{sec:punct}).
We use \punct to test whether a stronger separation between syntax and semantics could simplify the language generation task and ultimately translate to greater downstream performance.

\subsection{Vocabulary Size}

Vocabulary size, or number of tokens, is a key hyper-parameter that impacts the cost-efficiency of a Transformer model. 
While a larger vocabulary increases the cost per decoding step, it reduces both the memory needed for attention cache and the computation for generating a sentence. 
This increase in cost primarily affects the embedding and output layers. 
Therefore, in larger LLMs, the relative impact of a larger vocabulary on the overall parameter count becomes negligible (see Appendix~\ref{sec:vocab-size}). 
Consequently, for sufficiently large models, the benefits of a larger vocabulary, in terms of reduced total compute and memory requirements at inference, can outweigh the costs.

Large vocabulary sizes could also have adverse effects on downstream performance: with a large vocabulary every token is seen less on average by the model. 
This is a natural consequence of Zipf's law \cite{zipf1949human}. 
Therefore, we test whether tokenizer vocabulary size impacts performance in Section~\ref{sec:vocab-size-influence}.

\subsubsection{Optimal Vocabulary Size}
\label{sec:optimal-size}

Figure~\ref{fig:parameter-optimal} (top left) shows the Code NSL curve for a tokenizers trained on the same set of data, with varying vocabulary sizes from 10k to 256k.
We normalize the plot to a vocabulary size of $32,000$ (Code NSL@32k).
The gains in compression exponentially decrease as the vocabulary size increases, which indicates that there is an optimum point for a given downstream application where the additional token is not worth its added cost in compute or memory.

\textbf{Inference Optimal} We run experiments with different model sizes to measure to the effects of the vocabulary size on inference time with a fixed sequence length.
In Figure~\ref{fig:parameter-optimal} (top middle), we show these observations and plot the linear regressions found for each LLM size.
We normalize the observations and predictions to a vocabulary size of 32k.
Since the NSL describes the length of a sequence, it directly affects inference time.
We thus calculate the trade-off cost as the product between NSL@32k (top left) and the normalized inference time for each vocabulary size (top middle).
We plot these trade-off curves and find the minimum point for each LLM size in Figure~\ref{fig:parameter-optimal} (top right).
We find optimal inference time vocabulary size to grow with the size of the LLM. 
For LLMs like \llama~30B, we prefer even small gains in tokenizer compression, despite additional tokens in the final softmax, because of the high base cost of the forward pass.
Note that these calculations are heavily dependent on both hardware and software optimizations.

\textbf{Memory Optimal} The memory costs of running a LLM at inference time are mostly due to the weights of the model itself and its attention cache. At inference time, if the input and output matrices are not shared, the impact of increasing the size of the vocabulary $v$ on the number of additional parameters is:
\begin{equation}
    M(v) = 2 * dim * v
\end{equation}

Recall that sequence size is affected by the compression of the tokenizer, which is itself affected by the vocabulary size.
We can use the NSL@32k metric calculate the compressed length of a code sequence $s$ given a vocabulary size $v$ and the length of a sequence $l_{32k}$ encoded at 32k:

\begin{equation}
    s(v) = l_{32k} * NSL@32k
\end{equation}

Therefore the effects of vocabulary size $v$ on the number of parameters to keep in the cache depends on the batch size $b$, number of layers $n$, the hidden dimension $dim$, the number of kv heads and the sequence length $s(v)$:
\begin{align}
C(v) = 2 \times n \times b \times dim \times \frac{n\_kv\_heads}{n\_heads} \times s(v)
\end{align}
\llama~2 34B and 70B use grouped-query attention~\citep{ainslie2023gqa}, with only 8 kv heads, which significantly reduce the number of parameters kept in the cache compared to the 7B and 13B versions.
We also assume model and cache parameters to be stored at the same precision level.
We calculate the total effect of $v$ on memory $T$ as:
\begin{equation}\label{eq:eq1}
T(v) = M(v) + C(v) \\
\end{equation}

Figure~\ref{fig:parameter-optimal} (bottom) shows the memory optimal vocabulary sizes for models under different sequence lengths and batch sizes.
For short sequences ($l_{32k}=1000$) and small batches ($b=1$) the gains in compression from expanding the vocabulary is not worth the additional memory costs it incurs.
However, when using longer sequences or larger batches, the memory savings from the cache $C(v)$ are worth the additional memory costs $M(v)$ of an expanding vocabulary.

\section{Code Tokenizers Experiments}
\label{sec:experiments}

\begin{table}[h]
\scalebox{0.65}{
\begin{tabular}{lr|r|r|r|r|r}
\toprule 
regexp & \multicolumn{3}{c}{\textbf{HumanEval}} & \multicolumn{3}{c}{\textbf{MBPP}} \\
 & \multicolumn{1}{l}{\small{\textbf{Pass@1}}} & \multicolumn{1}{l}{\small{\textbf{Pass@100}}} & \multicolumn{1}{l}{\small{\textbf{Compile@1}}} & \multicolumn{1}{l}{\small{\textbf{Pass@1}}} & \multicolumn{1}{l}{\small{\textbf{Pass@100}}} & \multicolumn{1}{l}{\small{\textbf{Compile@1}}} \\
\toprule 
\small{\identitytable}\textsuperscript{R} & \acc{15.82621951}&\acc{49.90647628}& \acc{96.50609756}& \acc{21.72} & \acc{66.69038281} & \acc{98.967}\\
\small{\gpt}\textsuperscript{R} & \acc{17.80182927} & \acc{54.6479909} & \acc{98.967} & \acc{25.229} & \acc{70.71734375} & \acc{99.119}\\
\small{\identitytable}\textsuperscript{NL} & \acc{17.95731707} & \acc{60.0986209} & \acc{97.4695122} & \acc{24.544} & \acc{67.19198438} & \acc{99.055}\\
\small{\llama}\textsuperscript{NL} & \acc{20.5} & \acc{66.1} & \acc{98.4} & \acc{28.0} & \acc{71.2} & \acc{99.4} \\
\small{\puncttable}\textsuperscript{NL} & \acc{21.1} & \acc{63.5} & \acc{97.6} & \acc{28.6} & \acc{72.6} & \acc{99.5} \\
\small{\gpt}\textsuperscript{NL} & \acc{20.5} & \acc{65.3} & \acc{98.0} & \acc{27.2} & \acc{70.8} & \acc{99.4} \\
\small{\identitytable}\textsuperscript{C} & \acc{18.8} & \acc{66.1} & \acc{96.7} & \acc{23.4} & \acc{67.8} & \acc{99.4} \\
\small{\llama}\textsuperscript{C} & \acc{22.8} & \acc{68.6} & \acc{98.8} & \acc{30.1} & \acc{74.4} & \acc{99.4} \\
\small{\puncttable}\textsuperscript{C} & \acc{22.2} & \acc{68.7} & \acc{98.1} & \acc{29.8} & \acc{73.6} & \acc{99.4} \\
\small{\gpt}\textsuperscript{C} & \acc{21.2} & \acc{67.5} & \acc{97.7} & \acc{28.9} & \acc{74.9} & \acc{99.4} \\
\bottomrule
\end{tabular}
}
\caption{
We report the performance of fine-tuned 1.5B models using different tokenizers and base models on our two task generation tasks (HumanEval and MBPP) after 500B tokens seen. 
All tokenizers presented here are of vocabulary size 32k -- see Table~\ref{tab:NSL} for compression statistics for each. \textsuperscript{R} indicates a random initialization, \textsuperscript{NL} indicates that the base model used is our \smallmodel~1.5B model, and \textsuperscript{C} indicates that the base model is \smallcodemodel~1.5B. 
}
\label{tab:finetuning-32k}
\end{table}

\begin{figure}
    \centering
    \includegraphics[width=\linewidth]{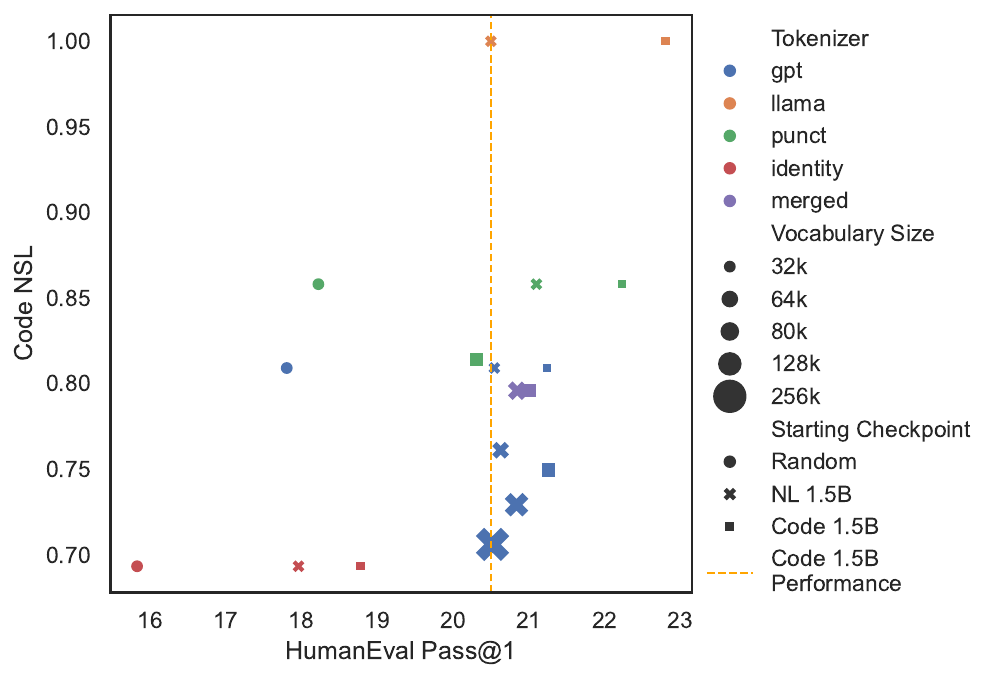}
    \caption{\textbf{Performance vs Code NSL}. We plot the HumanEval Pass@1 performance against Code NSL for our 1.5B LLMs fine-tuned with different base models and tokenizers.}
    \label{fig:performance-vs-nsl}
\end{figure}

We ask the question: \textit{what would have happened if \codellama had switched its tokenizer during fine-tuning}?
To answer this question, we first train two GPT-2 XL 1.5B base models \cite{gpt2} with the \llama tokenizer which we refer to as \smallmodel~1.5B and \smallcodemodel~1.5B respectively. 
\smallmodel~1.5B is trained for 1T tokens on a general data-mix resembling that of \llama~2 and \smallcodemodel~1.5B is further fine-tuned from \smallmodel~1.5B for 500B tokens using a code-specific data-mix similar to that of \codellama (see Appendix~\ref{sec:base-models} for more details).

We use the HumanEval \cite{humaneval} and MBPP \cite{mbpp} datasets to test the downstream effect of tokenizer change on LLMs.
For all evaluations, we generate $n=200$ samples per example with a temperature of $0.6$ and $top\_p=0.95$, and use the unbiased estimator detailed in \citet{humaneval} to calculate Pass@1, Pass@100, and Compile@1.
The Pass@k score~\cite{kulal2019spoc,roziere2020unsupervised,humaneval} measures the semantic correctness of a code snippet, by checking whether at least one of the k generated solution passes \textbf{all} the available unit tests.
Similarly, the Compile@k metric measures whether at least one of the k generated solutions compiles.
Note that we also use \textit{token healing} for all generations, which is a simple decoding trick to align the prompt along token boundaries (see Appendix~\ref{sec:token-healing}).

We report fine-tuning \smallmodel~1.5B and \smallcodemodel~1.5B, changing the tokenizer to our tokenizers of vocabulary size 32k (see Table~\ref{tab:NSL} for compression statistics) or keeping the \llama tokenizer constant.
See Table~\ref{tab:finetuning-32k} for the downstream performance on code generation of different base model/tokenizer configurations after fine-tuning.
For all tokenizer fine-tuning, whenever applicable, we apply Fast Vocabulary Transfer (FVT) \cite{FVT} to initialize the weights of the new embeddings.

The tokenizer compression impacts how much data the model sees for a given number of tokens (the amount of compute). 
As compute is often the primary constraint in training LLMs, we believe that assessing the \textit{token-equivalent} downstream performance, measured after training with the same number of tokens (500B), offers a fairer comparison between models. 
However, we also report \textit{word-equivalent} performance in Appendix~\ref{sec:word-equivalent}, which compares models trained on the same number of characters.

Table~\ref{tab:finetuning-32k} shows that when training a LLM from scratch on the same dataset (~\textsuperscript{R}), we obtain worse performance than if we use \smallmodel as the base model (~\textsuperscript{NL}).
This shows that the pre-trained model weights from the base model are still leveraged after a tokenizer change.
We see that this is the case as well when starting from a \codellama type model (~\textsuperscript{C}), where we get the best performance on the end-task when using \smallcodemodel~1.5B as the base model regardless of tokenizer.
\llama\textsuperscript{NL} is analogous to the original \codellama by \citet{roziere2023code} -- a \llama base model trained for 500B tokens on code without changing the tokenizer. 
\llama\textsuperscript{C} would be that same \codellama model fine-tuned for 500B more code tokens, and indicates further fine-tuning can still provide gains in code-generation performance.

While the \identity tokenizer has the greatest compression out of the tokenizers evaluated, it results in clear deteriorated performance on downstream code generation tasks.
Compared to the \llama tokenizer, the 32k \identity model compresses code 30\% more efficiently but its downstream performance is significantly worse on all metrics.
Moreover, forgoing token healing (see Appendix~\ref{sec:token-healing}) is especially detrimental with the \identity tokenizer.

In terms of differences between the other tokenizers, we see in Table~\ref{tab:finetuning-32k} that, at a vocabulary size of 32k, both \punct\textsuperscript{NL} and \gpt\textsuperscript{NL} obtain similar performance as \llama\textsuperscript{NL}.
On some metrics, such as Pass@1, \punct\textsuperscript{NL} even surpasses the \llama\textsuperscript{NL} baseline.
This is surprising, because it indicates that we can obtain both better performance and better compression by changing the tokenizer of a pre-trained LLM.
Therefore, we conclude that \textit{if \codellama had changed its tokenizer before fine-tuning, it would have had a negligible impact on downstream performance, but a large positive impact on compression and inference speed}.

\subsection{How much data?} 
\begin{figure}
    \centering
    \scalebox{0.5}{
    \includegraphics{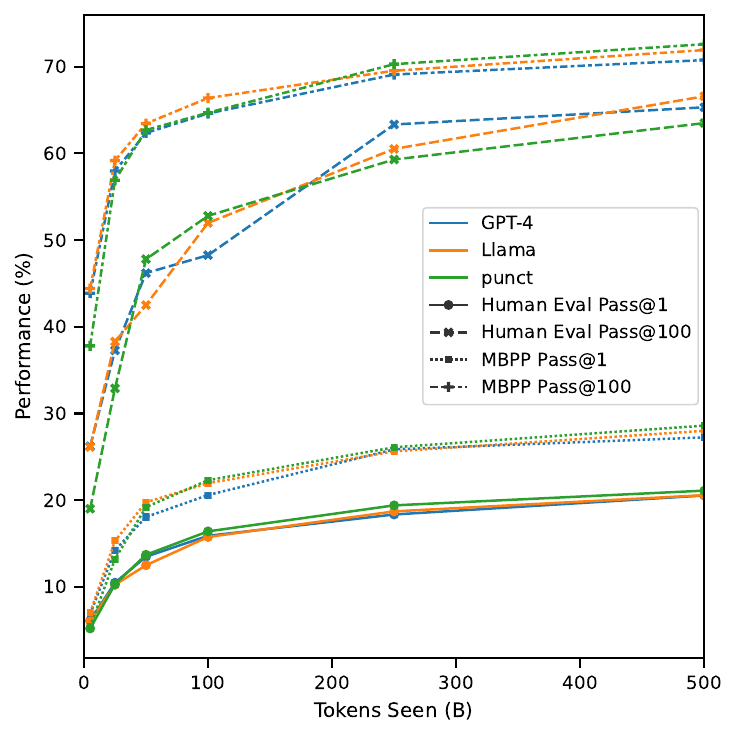}
    }
    \caption{Performance of \gpt\textsuperscript{NL}, \punct\textsuperscript{NL} and \llama\textsuperscript{NL} 1.5B at varying dataset sizes (5B, 25B, 50B, 100B, 250B, 500B).
    This figure demonstrates the impact of the number of tokens seen during training on end-task performance for models where the tokenizer is changed.
    We find both \gpt\textsuperscript{NL} and \punct\textsuperscript{NL} competitive with a LLM where the tokenizer is kept the same (\llama\textsuperscript{NL}), after 50B tokens tokens seen during training.
    }
\label{fig:token-seen}
\end{figure}

It is clear that changing a tokenizer has large effect on the weights, and crucially that not every fine-tuning regime would be appropriate for a tokenizer change. 
Specifically, we want to measure how many tokens does it take for a LLM to recover performance on end-tasks after suffering from a switch in tokenizer.

We fine-tune the \gpt\textsuperscript{NL}, \punct\textsuperscript{NL} and \llama\textsuperscript{NL} 32k 1.5B models on different subset sizes of the data ($5$B, $25$B, $50$B , $100$B, $250$B, $500$B) and measure downstream performance.
We keep all hyper-parameters and only adjust the learning rate schedule to fit the reduced datasets.
We report our results in Figure~\ref{fig:token-seen}.
While we find a large difference between \punct\textsuperscript{NL} and \llama\textsuperscript{NL} when trained on only 5B tokens, this difference almost disappears and even inverts (on Pass@1) after 50B tokens.
We therefore can only recommend that tokenizers are fine-tuned in regimes where there is enough training data for the model to adapt to the new distribution.

\subsection{Influence of tokenizer size}
\label{sec:vocab-size-influence}

\begin{table}[h]
\centering
\scalebox{0.85}{
\begin{tabular}{lrrrr}
\toprule
& \multicolumn{2}{c}{\textbf{Human Eval}} & \multicolumn{2}{c}{\textbf{MBPP}} \\
 & \multicolumn{1}{l}{\textbf{Pass@1}} & \multicolumn{1}{l}{\textbf{Pass@100}} & \multicolumn{1}{l}{\textbf{Pass@1}} & \multicolumn{1}{l}{\textbf{Pass@100}} \\
\midrule
32k & \acc{20.54268293} & \acc{65.32398652} & \acc{27.245} & \acc{70.77008594} \\
64k & \acc{20.625} & \acc{67.23258861} & \acc{27.603} & \acc{70.31582031} \\
128k & \acc{20.83536585} & \acc{64.22194169} & \acc{27.459} & \acc{70.821625} \\
256k & \acc{20.51829268} & \acc{63.53722251} & \acc{27.6} & \acc{71.24942969} \\
\bottomrule
\end{tabular}
}
\caption{
\textbf{Downstream performance \gpt\textsuperscript{NL} 1.5B model depending on tokenizer size.}
}
\label{tab:tokenizer-size}
\end{table}

We test the hypothesis of whether larger vocabulary sizes decrease downstream performance. 
As shown in Table~\ref{tab:tokenizer-size}, we change the tokenizer of a \smallmodel~1.5B model to that of a \gpt tokenizer of sizes 32k, 64k, 128k, and 256k.
Calculating the Pearson correlation coefficient between the tokenizer vocabulary size and HumanEval Pass@1, results in a correlation coefficient of $-0.13$ with a p-value of $0.87$.
This indicates a very weak inverse relationship between vocabulary size and HumanEval Pass@1, but the high p-value indicates that this correlation is far from statistically significant.
Therefore, we do not find that the vocabulary size (at this magnitude) to have an impact on end-goal performance and \textit{reject the hypothesis that larger vocabulary sizes decrease task performance}.

\subsection{Tokenizer update methods}
In this section, we compare two methods to update the tokenizer of a pre-trained model: using Fast Vocabulary Transfer (FVT) and extending an existing tokenizer. In Appendix~\ref{freezing_weights}, we also experiment with updating only the embedding and output weights, however this does not lead to improvements over full fine-tuning.

\begin{table}[h]
\scalebox{0.85}{
\begin{tabular}{lrrrr}
\toprule
& \multicolumn{2}{c}{\textbf{Human Eval}} & \multicolumn{2}{c}{\textbf{MBPP}} \\
 & \multicolumn{1}{l}{\textbf{Pass@1}} & \multicolumn{1}{l}{\textbf{Pass@100}} & \multicolumn{1}{l}{\textbf{Pass@1}} & \multicolumn{1}{l}{\textbf{Pass@100}} \\
\midrule
\small{\gpt\textsuperscript{NL}} & \acc{20.54268293}& \acc{65.32398652} &  \acc{27.245}& \acc{70.77008594} \\ 
\small{No-FVT \gpt\textsuperscript{NL}} & \acc{18.40548780487805} & \acc{58.61249523628049} & \acc{25.001} & \acc{69.248046875} \\
\small{\mergedtable\textsuperscript{NL}} & \acc{20.84451219512195} & \acc{67.47689000571647} & \acc{27.589} & \acc{70.90303125}\\
\bottomrule
\end{tabular}
}
\caption{We compare the performance of \gpt\textsuperscript{NL} 32k with FVT and without FVT (No-FVT), and that of the extended \llama tokenizer (\merged).}
\label{tab:additional-exps}
\end{table}

\subsubsection{Vocabulary Transfer}

Techniques such as Vocabulary Initialization with Partial Inheritance (VIPI)~\cite{vipi} and Fast Vocabulary Transfer (FVT)~\cite{FVT} have been proposed to adapt the embedding space of pre-trained models by mapping a new tokenizer onto an existing vocabulary.
VIPI and FVT both act as a simple way to map the old embedding space onto the new, but still require a fine-tuning stage to align the new representations with the model.
\citet{FVT} adapt a BERT base model \cite{bert} with FVT using in-domain (medical, legal, and news) tokenizers to obtain efficiency gains at little cost to performance.
Since, we use FVT to initialize our embedding matrix, we ablate FVT in Table~\ref{tab:additional-exps}.
We confirm that FVT leads to noticeable improvement on all downstream tasks.

\subsubsection{Tokenizer Extension}

As an alternative to FVT, we study extending a tokenizer (e.g. the \llama tokenizer) by adding domain-specific tokens. 
As the extended tokenizer contains all the tokens of the previous tokenizer, it may make the transition smoother and have less impact on the performance of the model.

We train a 64k vocabulary tokenizer using Sentencepiece on code data, changing the pre-tokenization scheme to allow tokens over tabs and new line characters, and filter out tokens disallowed by the \gpt regular expression.
We combine this in-domain tokenizer with the \llama tokenizer to obtain a tokenizer of size 80k which we refer to as \merged (see Table~\ref{tab:NSL} for compression statistics).
Table~\ref{tab:additional-exps} reports the \merged\textsuperscript{NL} results.
We observe only small gains from starting with the \merged tokenizer compared to starting from an entirely distinct tokenizer such as \gpt.

\subsection{7B models}

\begin{table}[h]
\scalebox{0.9}{
\begin{tabular}{lrrrr}
\toprule
& \multicolumn{2}{c}{\textbf{Human Eval}} & \multicolumn{2}{c}{\textbf{MBPP}} \\
 & \multicolumn{1}{l}{\textbf{Pass@1}} & \multicolumn{1}{l}{\textbf{Pass@100}} & \multicolumn{1}{l}{\textbf{Pass@1}} & \multicolumn{1}{l}{\textbf{Pass@100}} \\
\midrule
\codellama & \acc{32.05792683} & \acc{84.24667135} & \acc{41.926} & \acc{81.815875} \\
\puncttable & \acc{30.40243902} & \acc{85.14930688} & \acc{41.436} & \acc{81.92208594} \\
\gpt & \acc{30.82317073} & \acc{86.19894484} & \acc{42.594} & \acc{83.41950781} \\
\mergedtable & \acc{31.22865854} & \acc{86.34163014} & \acc{42.008} & \acc{80.85007031} \\
\bottomrule
\end{tabular}
}
\caption{\textbf{Downstream performance of fine-tuned \llama~2 7Bs}. We report the Pass@1 performance on HumanEval and MBPP of three \llama~2 7B models after 500B additional tokens of code pre-training. We also report the \codellama~7B model as a baseline.
}
\label{tab:7b}
\end{table}

Lastly, we fine-tune three 7B \llama models for 500B tokens and show that, as in the 1.5B LLMs, changing the tokenizer does not significantly impact code-generation performance.
Table~\ref{tab:7b} shows the result of changing the original \llama tokenizer in the \llama~2 7B model from \citet{touvron2023llama2} to \punct, \gpt, and \merged tokenizers of vocabulary size 80k.
Note that we opt to test for a larger vocabulary size as we showed in Section~\ref{sec:optimal-size} that larger models such as a 7B can trade-off additional parameters for increased code compression.
We compare our models to the 7B \codellama model from \citet{roziere2023code}, which used the tokenizer of \llama.
Our results support our thesis that, with long-enough fine-tuning, tokenizers can be changed without sacrificing performance.

\section{Related Works}

\textbf{Tokenizers in Machine Translation}
Tokenization has been a significant area of interest in multilingual tasks, such as Machine Translation, as different tokenization schemes can markedly influence model performance.
\citet{sennrich-etal-2016-neural} introduced BPE for tokenization and were the first to use sub-word tokenization a solution for encoding rare or unseen words at test time.
\citet{rust-etal-2021-good} analyze the implications of multilingual tokenization schemes and find that specialized monolingual tokenizers outperform multilingual versions.
More recently, \citet{Liang2023XLMVOT} demonstrate that expanding tokenizer vocabulary size and allocating a specific token quota for each language can substantially enhance performance in large-scale multilingual tasks.
\citet{Liang2023XLMVOT} also argue that increasing the tokenizer size can be an efficient way to increase the number of trainable parameters, since only a fraction of the embedding matrix is used for a given input.

\textbf{Tokenizers in Code Generation}
In the field of code generation, most LLMs follow standard tokenizer hyper-parameters.
Models like SantaCoder~\citep{santacoder} and InCoder~\citep{incoder} train a 50k vocabulary tokenizer on code data.
Fine-tuned code models derived from natural language LLMs, such as Codex~\cite{humaneval}, CodeGeeX~\citep{CodeGeeX}, and CodeGen~\citep{Nijkamp_Pang_Hayashi_Tu_Wang_Zhou_Savarese_Xiong_2023}, do not update their tokenizers but extend existing ones like GPT-2’s\footnote{See Appendix~\ref{sec:tokenizer_reuse} for an overview of tokenizer re-use in code generation.}.
\codellama \cite{roziere2023code} fine-tunes \llama~2 \cite{touvron2023llama2} keeping the original 32k \llama tokenizer \cite{touvron2023llama}.
\citet{Chirkova_Troshin_2022} find that custom code-tokenization can compress sequences by up to $17\%$ without sacrificing performance on a PLBART~\cite{PLBART} model.

\section{Conclusion}

This paper presents a comprehensive analysis of the impact of tokenization in modern LLMs. 
Our study reveals that varying the size, pre-tokenization regular expressions, and training data of tokenizers can significantly impact the compression rate of the tokenizer.
We showed that BPE tokenizers can compress code sequences by more than 40\% compared to the \llama tokenizer, and more than 25\% without any performance degradation (using \gpt 256k).
On top of that, we show that we can modify the tokenizer of a pre-trained LLM during fine-tuning if training for long enough ($>50$B tokens).
For models such as \codellama, this allows for substantial gains in generation speed and effective context size, at no cost in performance.

We present recommendations to train efficient tokenizers, evaluating the effects of data distribution, vocabulary size and pre-tokenization scheme on the compression rates of tokenizers.
We find that vocabulary size has little impact on coding performance, and present methods to find an optimal vocabulary size to optimize either memory or inference speed.
For most use cases, we validate the use of the \gpt pre-tokenization regular expression, which strikes a good balance between compression and performance. 
Its downstream performance is similar to that of the \llama tokenizer, and close to \punct on coding benchmarks while bringing clear improvements in compression. 
We find that skipping pre-tokenization (\identity) can maximize compression at significant cost to performance.
However, methods involving large-scale sampling and more concerned with the pass@100 than pass@1, or requiring large context sizes, might make good use of the maximal-compression \identity pre-tokenization.
More broadly, although \punct sometimes outperforms \gpt, we ultimately recommend using the \gpt as it provides an additional $5\%$ compression. 
Finally, we hope this our results pushes researchers and practitioners to think further about the design of their tokenizer and consider changing to an in-domain tokenizer when fine-tuning.

\bibliography{custom, anthology}

\begin{thebibliography}{54}
\providecommand{\natexlab}[1]{#1}
\providecommand{\url}[1]{\texttt{#1}}
\expandafter\ifx\csname urlstyle\endcsname\relax
  \providecommand{\doi}[1]{doi: #1}\else
  \providecommand{\doi}{doi: \begingroup \urlstyle{rm}\Url}\fi

\bibitem[{01.AI}(2023)]{01ai_2023_yi6b}
{01.AI}.
\newblock Yi series models: Large language models.
\newblock Hugging Face Model Repository, 2023.
\newblock URL \url{https://huggingface.co/01-ai/Yi-6B}.
\newblock Accessed: 17/11/2023.

\bibitem[Ahmad et~al.(2021)Ahmad, Chakraborty, Ray, and Chang]{PLBART}
Ahmad, W., Chakraborty, S., Ray, B., and Chang, K.-W.
\newblock Unified pre-training for program understanding and generation.
\newblock In Toutanova, K., Rumshisky, A., Zettlemoyer, L., Hakkani-Tur, D.,
  Beltagy, I., Bethard, S., Cotterell, R., Chakraborty, T., and Zhou, Y.
  (eds.), \emph{Proceedings of the 2021 Conference of the North American
  Chapter of the Association for Computational Linguistics: Human Language
  Technologies}, pp.\  2655--2668, Online, June 2021. Association for
  Computational Linguistics.
\newblock \doi{10.18653/v1/2021.naacl-main.211}.
\newblock URL \url{https://aclanthology.org/2021.naacl-main.211}.

\bibitem[Ainslie et~al.(2023)Ainslie, Lee-Thorp, de~Jong, Zemlyanskiy, Lebron,
  and Sanghai]{ainslie2023gqa}
Ainslie, J., Lee-Thorp, J., de~Jong, M., Zemlyanskiy, Y., Lebron, F., and
  Sanghai, S.
\newblock {GQA}: Training generalized multi-query transformer models from
  multi-head checkpoints.
\newblock In Bouamor, H., Pino, J., and Bali, K. (eds.), \emph{Proceedings of
  the 2023 Conference on Empirical Methods in Natural Language Processing},
  pp.\  4895--4901, Singapore, December 2023. Association for Computational
  Linguistics.
\newblock \doi{10.18653/v1/2023.emnlp-main.298}.
\newblock URL \url{https://aclanthology.org/2023.emnlp-main.298}.

\bibitem[Allal et~al.(2023)Allal, Li, Kocetkov, Mou, Akiki, Ferrandis,
  Muennighoff, Mishra, Gu, Dey, Umapathi, Anderson, Zi, Lamy{-}Poirier,
  Schoelkopf, Troshin, Abulkhanov, Romero, Lappert, Toni, del R{\'{\i}}o, Liu,
  Bose, Bhattacharyya, Zhuo, Yu, Villegas, Zocca, Mangrulkar, Lansky, Nguyen,
  Contractor, Villa, Li, Bahdanau, Jernite, Hughes, Fried, Guha, de~Vries, and
  von Werra]{santacoder}
Allal, L.~B., Li, R., Kocetkov, D., Mou, C., Akiki, C., Ferrandis, C.~M.,
  Muennighoff, N., Mishra, M., Gu, A., Dey, M., Umapathi, L.~K., Anderson,
  C.~J., Zi, Y., Lamy{-}Poirier, J., Schoelkopf, H., Troshin, S., Abulkhanov,
  D., Romero, M., Lappert, M., Toni, F.~D., del R{\'{\i}}o, B.~G., Liu, Q.,
  Bose, S., Bhattacharyya, U., Zhuo, T.~Y., Yu, I., Villegas, P., Zocca, M.,
  Mangrulkar, S., Lansky, D., Nguyen, H., Contractor, D., Villa, L., Li, J.,
  Bahdanau, D., Jernite, Y., Hughes, S., Fried, D., Guha, A., de~Vries, H., and
  von Werra, L.
\newblock Santacoder: don't reach for the stars!
\newblock \emph{CoRR}, abs/2301.03988, 2023.
\newblock \doi{10.48550/ARXIV.2301.03988}.
\newblock URL \url{https://doi.org/10.48550/arXiv.2301.03988}.

\bibitem[Almazrouei et~al.(2023)Almazrouei, Alobeidli, Alshamsi, Cappelli,
  Cojocaru, Debbah, Étienne Goffinet, Hesslow, Launay, Malartic, Mazzotta,
  Noune, Pannier, and Penedo]{falcon40b}
Almazrouei, E., Alobeidli, H., Alshamsi, A., Cappelli, A., Cojocaru, R.,
  Debbah, M., Étienne Goffinet, Hesslow, D., Launay, J., Malartic, Q.,
  Mazzotta, D., Noune, B., Pannier, B., and Penedo, G.
\newblock The falcon series of open language models, 2023.

\bibitem[Anthropic(2023)]{anthropic_2023_claude}
Anthropic.
\newblock Introducing claude.
\newblock Anthropic Blog, March 2023.
\newblock URL \url{https://www.anthropic.com/index/introducing-claude}.
\newblock Accessed: 17/11/2023.

\bibitem[Austin et~al.(2021)Austin, Odena, Nye, Bosma, Michalewski, Dohan,
  Jiang, Cai, Terry, Le, et~al.]{mbpp}
Austin, J., Odena, A., Nye, M., Bosma, M., Michalewski, H., Dohan, D., Jiang,
  E., Cai, C., Terry, M., Le, Q., et~al.
\newblock Program synthesis with large language models.
\newblock \emph{arXiv preprint arXiv:2108.07732}, 2021.

\bibitem[Biderman et~al.(2023)Biderman, Schoelkopf, Anthony, Bradley,
  O’Brien, Hallahan, Khan, Purohit, Prashanth, Raff,
  et~al.]{Biderman2023PythiaAS}
Biderman, S., Schoelkopf, H., Anthony, Q.~G., Bradley, H., O’Brien, K.,
  Hallahan, E., Khan, M.~A., Purohit, S., Prashanth, U.~S., Raff, E., et~al.
\newblock Pythia: A suite for analyzing large language models across training
  and scaling.
\newblock In \emph{International Conference on Machine Learning}, pp.\
  2397--2430. PMLR, 2023.

\bibitem[Black et~al.(2022)Black, Biderman, Hallahan, Anthony, Gao, Golding,
  He, Leahy, McDonell, Phang, Pieler, Prashanth, Purohit, Reynolds, Tow, Wang,
  and Weinbach]{Black2022GPTNeoX20BAO}
Black, S., Biderman, S., Hallahan, E., Anthony, Q., Gao, L., Golding, L., He,
  H., Leahy, C., McDonell, K., Phang, J., Pieler, M., Prashanth, U.~S.,
  Purohit, S., Reynolds, L., Tow, J., Wang, B., and Weinbach, S.
\newblock {GPT}-{N}eo{X}-20{B}: An open-source autoregressive language model.
\newblock In Fan, A., Ilic, S., Wolf, T., and Gall{\'e}, M. (eds.),
  \emph{Proceedings of BigScience Episode {\#}5 -- Workshop on Challenges {\&}
  Perspectives in Creating Large Language Models}, pp.\  95--136,
  virtual+Dublin, May 2022. Association for Computational Linguistics.
\newblock \doi{10.18653/v1/2022.bigscience-1.9}.
\newblock URL \url{https://aclanthology.org/2022.bigscience-1.9}.

\bibitem[Chen et~al.(2021)Chen, Tworek, Jun, Yuan, de~Oliveira~Pinto, Kaplan,
  Edwards, Burda, Joseph, Brockman, Ray, Puri, Krueger, Petrov, Khlaaf, Sastry,
  Mishkin, Chan, Gray, Ryder, Pavlov, Power, Kaiser, Bavarian, Winter, Tillet,
  Such, Cummings, Plappert, Chantzis, Barnes, Herbert{-}Voss, Guss, Nichol,
  Paino, Tezak, Tang, Babuschkin, Balaji, Jain, Saunders, Hesse, Carr, Leike,
  Achiam, Misra, Morikawa, Radford, Knight, Brundage, Murati, Mayer, Welinder,
  McGrew, Amodei, McCandlish, Sutskever, and Zaremba]{humaneval}
Chen, M., Tworek, J., Jun, H., Yuan, Q., de~Oliveira~Pinto, H.~P., Kaplan, J.,
  Edwards, H., Burda, Y., Joseph, N., Brockman, G., Ray, A., Puri, R., Krueger,
  G., Petrov, M., Khlaaf, H., Sastry, G., Mishkin, P., Chan, B., Gray, S.,
  Ryder, N., Pavlov, M., Power, A., Kaiser, L., Bavarian, M., Winter, C.,
  Tillet, P., Such, F.~P., Cummings, D., Plappert, M., Chantzis, F., Barnes,
  E., Herbert{-}Voss, A., Guss, W.~H., Nichol, A., Paino, A., Tezak, N., Tang,
  J., Babuschkin, I., Balaji, S., Jain, S., Saunders, W., Hesse, C., Carr,
  A.~N., Leike, J., Achiam, J., Misra, V., Morikawa, E., Radford, A., Knight,
  M., Brundage, M., Murati, M., Mayer, K., Welinder, P., McGrew, B., Amodei,
  D., McCandlish, S., Sutskever, I., and Zaremba, W.
\newblock Evaluating large language models trained on code.
\newblock \emph{CoRR}, abs/2107.03374, 2021.
\newblock URL \url{https://arxiv.org/abs/2107.03374}.

\bibitem[Chirkova \& Troshin(2023)Chirkova and Troshin]{Chirkova_Troshin_2022}
Chirkova, N. and Troshin, S.
\newblock Codebpe: Investigating subtokenization options for large language
  model pretraining on source code.
\newblock In \emph{The Eleventh International Conference on Learning
  Representations, {ICLR} 2023, Kigali, Rwanda, May 1-5, 2023}. OpenReview.net,
  2023.
\newblock URL \url{https://openreview.net/pdf?id=htL4UZ344nF}.

\bibitem[{Deci}(2023)]{decicoder}
{Deci}.
\newblock Introducing decicoder: The new gold standard in efficient and
  accurate code generation, August 2023.
\newblock URL \url{https://deci.ai/blog/}.
\newblock Accessed: 2024-01-18.

\bibitem[{DeepSeek AI}(2023)]{deepseekai_2023_deepseekcoder13binstruct}
{DeepSeek AI}.
\newblock Deepseek coder: A series of code language models.
\newblock DeepSeek Coder Official Website, 2023.
\newblock URL \url{https://deepseekcoder.github.io/}.
\newblock Accessed: 17/11/2023.

\bibitem[Devlin et~al.(2019)Devlin, Chang, Lee, and Toutanova]{bert}
Devlin, J., Chang, M.-W., Lee, K., and Toutanova, K.
\newblock {BERT}: Pre-training of deep bidirectional transformers for language
  understanding.
\newblock In Burstein, J., Doran, C., and Solorio, T. (eds.), \emph{Proceedings
  of the 2019 Conference of the North {A}merican Chapter of the Association for
  Computational Linguistics: Human Language Technologies, Volume 1 (Long and
  Short Papers)}, pp.\  4171--4186, Minneapolis, Minnesota, June 2019.
  Association for Computational Linguistics.
\newblock \doi{10.18653/v1/N19-1423}.
\newblock URL \url{https://aclanthology.org/N19-1423}.

\bibitem[Elsen et~al.(2023)Elsen, Odena, Nye, Ta\c{s}\i{}rlar, Dao, Hawthorne,
  Moparthi, and Somani]{persimmon-8b}
Elsen, E., Odena, A., Nye, M., Ta\c{s}\i{}rlar, S., Dao, T., Hawthorne, C.,
  Moparthi, D., and Somani, A.
\newblock Releasing {Persimmon-8B}, 2023.
\newblock URL \url{https://www.adept.ai/blog/persimmon-8b}.

\bibitem[Forsythe(2023)]{tokenmonster}
Forsythe, A.
\newblock Tokenmonster: Ungreedy subword tokenizer and vocabulary trainer for
  python, go and javascript.
\newblock \url{https://github.com/alasdairforsythe/tokenmonster}, 2023.

\bibitem[Fried et~al.(2023)Fried, Aghajanyan, Lin, Wang, Wallace, Shi, Zhong,
  Yih, Zettlemoyer, and Lewis]{incoder}
Fried, D., Aghajanyan, A., Lin, J., Wang, S., Wallace, E., Shi, F., Zhong, R.,
  Yih, S., Zettlemoyer, L., and Lewis, M.
\newblock Incoder: {A} generative model for code infilling and synthesis.
\newblock In \emph{The Eleventh International Conference on Learning
  Representations, {ICLR} 2023, Kigali, Rwanda, May 1-5, 2023}. OpenReview.net,
  2023.
\newblock URL \url{https://openreview.net/pdf?id=hQwb-lbM6EL}.

\bibitem[Gee et~al.(2022)Gee, Zugarini, Rigutini, and Torroni]{FVT}
Gee, L., Zugarini, A., Rigutini, L., and Torroni, P.
\newblock Fast vocabulary transfer for language model compression.
\newblock In \emph{Proceedings of the 2022 Conference on Empirical Methods in
  Natural Language Processing: Industry Track}, pp.\  409--416, Abu Dhabi, UAE,
  December 2022. Association for Computational Linguistics.
\newblock \doi{10.18653/v1/2022.emnlp-industry.41}.
\newblock URL \url{https://aclanthology.org/2022.emnlp-industry.41}.

\bibitem[Gee et~al.(2023)Gee, Rigutini, Ernandes, and
  Zugarini]{gee-etal-2023-multi}
Gee, L., Rigutini, L., Ernandes, M., and Zugarini, A.
\newblock Multi-word tokenization for sequence compression.
\newblock In Wang, M. and Zitouni, I. (eds.), \emph{Proceedings of the 2023
  Conference on Empirical Methods in Natural Language Processing: Industry
  Track}, pp.\  612--621, Singapore, December 2023. Association for
  Computational Linguistics.
\newblock \doi{10.18653/v1/2023.emnlp-industry.58}.
\newblock URL \url{https://aclanthology.org/2023.emnlp-industry.58}.

\bibitem[Gowda \& May(2020)Gowda and May]{gowda2020finding}
Gowda, T. and May, J.
\newblock Finding the optimal vocabulary size for neural machine translation.
\newblock In \emph{Findings of the Association for Computational Linguistics:
  EMNLP 2020}, pp.\  3955–3964, Online, Nov 2020. Association for
  Computational Linguistics.
\newblock \doi{10.18653/v1/2020.findings-emnlp.352}.
\newblock URL \url{https://aclanthology.org/2020.findings-emnlp.352}.

\bibitem[Goyal et~al.(2023)Goyal, Ji, Rawat, Menon, Kumar, and
  Nagarajan]{pause_tokens}
Goyal, S., Ji, Z., Rawat, A.~S., Menon, A.~K., Kumar, S., and Nagarajan, V.
\newblock Think before you speak: Training language models with pause tokens.
\newblock \emph{CoRR}, abs/2310.02226, 2023.
\newblock \doi{10.48550/ARXIV.2310.02226}.
\newblock URL \url{https://doi.org/10.48550/arXiv.2310.02226}.

\bibitem[{guidance-ai}(2023)]{guidance_ai_guidance_2024}
{guidance-ai}.
\newblock {A guidance language for controlling large language models}.
\newblock \url{https://github.com/guidance-ai/guidance}, 2023.
\newblock Accessed: 2024-01-15.

\bibitem[Jiang et~al.(2023)Jiang, Sablayrolles, Mensch, Bamford, Chaplot,
  de~Las~Casas, Bressand, Lengyel, Lample, Saulnier, Lavaud, Lachaux, Stock,
  Scao, Lavril, Wang, Lacroix, and Sayed]{jiang2023mistral}
Jiang, A.~Q., Sablayrolles, A., Mensch, A., Bamford, C., Chaplot, D.~S.,
  de~Las~Casas, D., Bressand, F., Lengyel, G., Lample, G., Saulnier, L.,
  Lavaud, L.~R., Lachaux, M., Stock, P., Scao, T.~L., Lavril, T., Wang, T.,
  Lacroix, T., and Sayed, W.~E.
\newblock Mistral 7b.
\newblock \emph{CoRR}, abs/2310.06825, 2023.
\newblock \doi{10.48550/ARXIV.2310.06825}.
\newblock URL \url{https://doi.org/10.48550/arXiv.2310.06825}.

\bibitem[Kocetkov et~al.(2022)Kocetkov, Li, Allal, Li, Mou, Ferrandis, Jernite,
  Mitchell, Hughes, Wolf, Bahdanau, von Werra, and de~Vries]{Kocetkov2022TheS3}
Kocetkov, D., Li, R., Allal, L.~B., Li, J., Mou, C., Ferrandis, C.~M., Jernite,
  Y., Mitchell, M., Hughes, S., Wolf, T., Bahdanau, D., von Werra, L., and
  de~Vries, H.
\newblock The stack: 3 {TB} of permissively licensed source code.
\newblock \emph{CoRR}, abs/2211.15533, 2022.
\newblock \doi{10.48550/ARXIV.2211.15533}.
\newblock URL \url{https://doi.org/10.48550/arXiv.2211.15533}.

\bibitem[Kudo(2018)]{Kudo_2018}
Kudo, T.
\newblock Subword regularization: Improving neural network translation models
  with multiple subword candidates.
\newblock In Gurevych, I. and Miyao, Y. (eds.), \emph{Proceedings of the 56th
  Annual Meeting of the Association for Computational Linguistics, {ACL} 2018,
  Melbourne, Australia, July 15-20, 2018, Volume 1: Long Papers}, pp.\  66--75.
  Association for Computational Linguistics, 2018.
\newblock \doi{10.18653/V1/P18-1007}.
\newblock URL \url{https://aclanthology.org/P18-1007/}.

\bibitem[Kudo \& Richardson(2018)Kudo and Richardson]{Kudo2018SentencePieceAS}
Kudo, T. and Richardson, J.
\newblock Sentencepiece: {A} simple and language independent subword tokenizer
  and detokenizer for neural text processing.
\newblock In Blanco, E. and Lu, W. (eds.), \emph{Proceedings of the 2018
  Conference on Empirical Methods in Natural Language Processing, {EMNLP} 2018:
  System Demonstrations, Brussels, Belgium, October 31 - November 4, 2018},
  pp.\  66--71. Association for Computational Linguistics, 2018.
\newblock \doi{10.18653/V1/D18-2012}.
\newblock URL \url{https://doi.org/10.18653/v1/d18-2012}.

\bibitem[Kulal et~al.(2019)Kulal, Pasupat, Chandra, Lee, Padon, Aiken, and
  Liang]{kulal2019spoc}
Kulal, S., Pasupat, P., Chandra, K., Lee, M., Padon, O., Aiken, A., and Liang,
  P.
\newblock Spoc: Search-based pseudocode to code.
\newblock In Wallach, H.~M., Larochelle, H., Beygelzimer, A.,
  d'Alch{\'{e}}{-}Buc, F., Fox, E.~B., and Garnett, R. (eds.), \emph{Advances
  in Neural Information Processing Systems 32: Annual Conference on Neural
  Information Processing Systems 2019, NeurIPS 2019, December 8-14, 2019,
  Vancouver, BC, Canada}, pp.\  11883--11894, 2019.
\newblock URL
  \url{https://proceedings.neurips.cc/paper/2019/hash/7298332f04ac004a0ca44cc69ecf6f6b-Abstract.html}.

\bibitem[Li et~al.(2023)Li, Allal, Zi, Muennighoff, Kocetkov, Mou, Marone,
  Akiki, Li, Chim, Liu, Zheltonozhskii, Zhuo, Wang, Dehaene, Davaadorj,
  Lamy{-}Poirier, Monteiro, Shliazhko, Gontier, Meade, Zebaze, Yee, Umapathi,
  Zhu, Lipkin, Oblokulov, Wang, V, Stillerman, Patel, Abulkhanov, Zocca, Dey,
  Zhang, Moustafa{-}Fahmy, Bhattacharyya, Yu, Singh, Luccioni, Villegas,
  Kunakov, Zhdanov, Romero, Lee, Timor, Ding, Schlesinger, Schoelkopf, Ebert,
  Dao, Mishra, Gu, Robinson, Anderson, Dolan{-}Gavitt, Contractor, Reddy,
  Fried, Bahdanau, Jernite, Ferrandis, Hughes, Wolf, Guha, von Werra, and
  de~Vries]{starcoder}
Li, R., Allal, L.~B., Zi, Y., Muennighoff, N., Kocetkov, D., Mou, C., Marone,
  M., Akiki, C., Li, J., Chim, J., Liu, Q., Zheltonozhskii, E., Zhuo, T.~Y.,
  Wang, T., Dehaene, O., Davaadorj, M., Lamy{-}Poirier, J., Monteiro, J.,
  Shliazhko, O., Gontier, N., Meade, N., Zebaze, A., Yee, M., Umapathi, L.~K.,
  Zhu, J., Lipkin, B., Oblokulov, M., Wang, Z., V, R.~M., Stillerman, J.,
  Patel, S.~S., Abulkhanov, D., Zocca, M., Dey, M., Zhang, Z.,
  Moustafa{-}Fahmy, N., Bhattacharyya, U., Yu, W., Singh, S., Luccioni, S.,
  Villegas, P., Kunakov, M., Zhdanov, F., Romero, M., Lee, T., Timor, N., Ding,
  J., Schlesinger, C., Schoelkopf, H., Ebert, J., Dao, T., Mishra, M., Gu, A.,
  Robinson, J., Anderson, C.~J., Dolan{-}Gavitt, B., Contractor, D., Reddy, S.,
  Fried, D., Bahdanau, D., Jernite, Y., Ferrandis, C.~M., Hughes, S., Wolf, T.,
  Guha, A., von Werra, L., and de~Vries, H.
\newblock Starcoder: may the source be with you!
\newblock \emph{CoRR}, abs/2305.06161, 2023.
\newblock \doi{10.48550/ARXIV.2305.06161}.
\newblock URL \url{https://doi.org/10.48550/arXiv.2305.06161}.

\bibitem[Liang et~al.(2023)Liang, Gonen, Mao, Hou, Goyal, Ghazvininejad,
  Zettlemoyer, and Khabsa]{Liang2023XLMVOT}
Liang, D., Gonen, H., Mao, Y., Hou, R., Goyal, N., Ghazvininejad, M.,
  Zettlemoyer, L., and Khabsa, M.
\newblock {XLM-V:} overcoming the vocabulary bottleneck in multilingual masked
  language models.
\newblock In Bouamor, H., Pino, J., and Bali, K. (eds.), \emph{Proceedings of
  the 2023 Conference on Empirical Methods in Natural Language Processing,
  {EMNLP} 2023, Singapore, December 6-10, 2023}, pp.\  13142--13152.
  Association for Computational Linguistics, 2023.
\newblock URL \url{https://aclanthology.org/2023.emnlp-main.813}.

\bibitem[Limisiewicz et~al.(2023)Limisiewicz, Balhar, and
  Mare{\v{c}}ek]{limisiewicz-etal-2023-tokenization}
Limisiewicz, T., Balhar, J., and Mare{\v{c}}ek, D.
\newblock Tokenization impacts multilingual language modeling: Assessing
  vocabulary allocation and overlap across languages.
\newblock In Rogers, A., Boyd-Graber, J., and Okazaki, N. (eds.),
  \emph{Findings of the Association for Computational Linguistics: ACL 2023},
  pp.\  5661--5681, Toronto, Canada, July 2023. Association for Computational
  Linguistics.
\newblock \doi{10.18653/v1/2023.findings-acl.350}.
\newblock URL \url{https://aclanthology.org/2023.findings-acl.350}.

\bibitem[Loshchilov \& Hutter(2019)Loshchilov and
  Hutter]{Loshchilov2017DecoupledWD}
Loshchilov, I. and Hutter, F.
\newblock Decoupled weight decay regularization.
\newblock In \emph{7th International Conference on Learning Representations,
  {ICLR} 2019, New Orleans, LA, USA, May 6-9, 2019}. OpenReview.net, 2019.
\newblock URL \url{https://openreview.net/forum?id=Bkg6RiCqY7}.

\bibitem[{MosaicML}(2023)]{mpt}
{MosaicML}.
\newblock Mpt-7b-instruct: A model for short-form instruction following.
\newblock MosaicML Blog, May 2023.
\newblock URL \url{https://www.mosaicml.com/blog/mpt-7b}.
\newblock Accessed: 17/11/2023.

\bibitem[Mosin et~al.(2023)Mosin, Samenko, Kozlovskii, Tikhonov, and
  Yamshchikov]{vipi}
Mosin, V.~D., Samenko, I., Kozlovskii, B., Tikhonov, A., and Yamshchikov, I.~P.
\newblock Fine-tuning transformers: Vocabulary transfer.
\newblock \emph{Artif. Intell.}, 317:\penalty0 103860, 2023.
\newblock \doi{10.1016/J.ARTINT.2023.103860}.
\newblock URL \url{https://doi.org/10.1016/j.artint.2023.103860}.

\bibitem[Nijkamp et~al.(2023)Nijkamp, Pang, Hayashi, Tu, Wang, Zhou, Savarese,
  and Xiong]{Nijkamp_Pang_Hayashi_Tu_Wang_Zhou_Savarese_Xiong_2023}
Nijkamp, E., Pang, B., Hayashi, H., Tu, L., Wang, H., Zhou, Y., Savarese, S.,
  and Xiong, C.
\newblock Codegen: An open large language model for code with multi-turn
  program synthesis.
\newblock In \emph{The Eleventh International Conference on Learning
  Representations, {ICLR} 2023, Kigali, Rwanda, May 1-5, 2023}. OpenReview.net,
  2023.
\newblock URL \url{https://openreview.net/pdf?id=iaYcJKpY2B\_}.

\bibitem[Nogueira et~al.(2021)Nogueira, Jiang, and
  Lin]{nogueira2021investigating}
Nogueira, R.~F., Jiang, Z., and Lin, J.
\newblock Investigating the limitations of the transformers with simple
  arithmetic tasks.
\newblock \emph{CoRR}, abs/2102.13019, 2021.
\newblock URL \url{https://arxiv.org/abs/2102.13019}.

\bibitem[OpenAI(2023)]{openai2023gpt4}
OpenAI.
\newblock {GPT-4} technical report.
\newblock \emph{CoRR}, abs/2303.08774, 2023.
\newblock \doi{10.48550/ARXIV.2303.08774}.
\newblock URL \url{https://doi.org/10.48550/arXiv.2303.08774}.

\bibitem[Provilkov et~al.(2020)Provilkov, Emelianenko, and
  Voita]{Provilkov_Emelianenko_Voita_2020}
Provilkov, I., Emelianenko, D., and Voita, E.
\newblock Bpe-dropout: Simple and effective subword regularization.
\newblock In \emph{Proceedings of the 58th Annual Meeting of the Association
  for Computational Linguistics}, pp.\  1882–1892, Online, Jul 2020.
  Association for Computational Linguistics.
\newblock \doi{10.18653/v1/2020.acl-main.170}.
\newblock URL \url{https://aclanthology.org/2020.acl-main.170}.

\bibitem[Radford et~al.(2019)Radford, Wu, Child, Luan, Amodei, and
  Sutskever]{gpt2}
Radford, A., Wu, J., Child, R., Luan, D., Amodei, D., and Sutskever, I.
\newblock Language models are unsupervised multitask learners.
\newblock 2019.
\newblock URL \url{https://api.semanticscholar.org/CorpusID:160025533}.

\bibitem[{Replit}(2023)]{replit}
{Replit}.
\newblock Replit’s new ai model now available on hugging face.
\newblock Replit Blog, October 2023.
\newblock URL \url{https://blog.replit.com/replit-code-v1_5}.
\newblock Accessed: 17/11/2023.

\bibitem[Rozi{\`{e}}re et~al.(2020)Rozi{\`{e}}re, Lachaux, Chanussot, and
  Lample]{roziere2020unsupervised}
Rozi{\`{e}}re, B., Lachaux, M., Chanussot, L., and Lample, G.
\newblock Unsupervised translation of programming languages.
\newblock In Larochelle, H., Ranzato, M., Hadsell, R., Balcan, M., and Lin, H.
  (eds.), \emph{Advances in Neural Information Processing Systems 33: Annual
  Conference on Neural Information Processing Systems 2020, NeurIPS 2020,
  December 6-12, 2020, virtual}, 2020.
\newblock URL
  \url{https://proceedings.neurips.cc/paper/2020/hash/ed23fbf18c2cd35f8c7f8de44f85c08d-Abstract.html}.

\bibitem[Rozi{\`{e}}re et~al.(2023)Rozi{\`{e}}re, Gehring, Gloeckle, Sootla,
  Gat, Tan, Adi, Liu, Remez, Rapin, Kozhevnikov, Evtimov, Bitton, Bhatt,
  Canton{-}Ferrer, Grattafiori, Xiong, D{\'{e}}fossez, Copet, Azhar, Touvron,
  Martin, Usunier, Scialom, and Synnaeve]{roziere2023code}
Rozi{\`{e}}re, B., Gehring, J., Gloeckle, F., Sootla, S., Gat, I., Tan, X.~E.,
  Adi, Y., Liu, J., Remez, T., Rapin, J., Kozhevnikov, A., Evtimov, I., Bitton,
  J., Bhatt, M., Canton{-}Ferrer, C., Grattafiori, A., Xiong, W.,
  D{\'{e}}fossez, A., Copet, J., Azhar, F., Touvron, H., Martin, L., Usunier,
  N., Scialom, T., and Synnaeve, G.
\newblock Code llama: Open foundation models for code.
\newblock \emph{CoRR}, abs/2308.12950, 2023.
\newblock \doi{10.48550/ARXIV.2308.12950}.
\newblock URL \url{https://doi.org/10.48550/arXiv.2308.12950}.

\bibitem[Rust et~al.(2021)Rust, Pfeiffer, Vuli{\'c}, Ruder, and
  Gurevych]{rust-etal-2021-good}
Rust, P., Pfeiffer, J., Vuli{\'c}, I., Ruder, S., and Gurevych, I.
\newblock How good is your tokenizer? on the monolingual performance of
  multilingual language models.
\newblock In \emph{Proceedings of the 59th Annual Meeting of the Association
  for Computational Linguistics and the 11th International Joint Conference on
  Natural Language Processing (Volume 1: Long Papers)}, pp.\  3118--3135,
  Online, August 2021. Association for Computational Linguistics.
\newblock \doi{10.18653/v1/2021.acl-long.243}.
\newblock URL \url{https://aclanthology.org/2021.acl-long.243}.

\bibitem[Scao et~al.(2022)Scao, Fan, Akiki, Pavlick, Ili'c, Hesslow, Castagn'e,
  Luccioni, Yvon, Gall{\'e}, Tow, Rush, Biderman, Webson, Ammanamanchi, Wang,
  Sagot, Muennighoff, del Moral, Ruwase, Bawden, Bekman, McMillan-Major,
  Beltagy, Nguyen, Saulnier, Tan, Suarez, Sanh, Laurenccon, Jernite, Launay,
  Mitchell, Raffel, Gokaslan, Simhi, Etxabe, Aji, Alfassy, Rogers, Nitzav, Xu,
  Mou, Emezue, Klamm, Leong, van Strien, Adelani, Radev, Ponferrada, Levkovizh,
  Kim, Natan, Toni, Dupont, Kruszewski, Pistilli, ElSahar, Benyamina, Tran, Yu,
  Abdulmumin, Johnson, Gonzalez-Dios, de~la Rosa, Chim, Dodge, Zhu, Chang,
  Frohberg, Tobing, Bhattacharjee, Almubarak, Chen, Lo, von Werra, Weber, Phan,
  Allal, Tanguy, Dey, Mu{\~n}oz, Masoud, Grandury, vSavsko, Huang, Coavoux,
  Singh, Jiang, Vu, Jauhar, Ghaleb, Subramani, Kassner, Khamis, Nguyen,
  Espejel, de~Gibert, Villegas, Henderson, Colombo, Amuok, Lhoest, Harliman,
  Bommasani, L'opez, Ribeiro, Osei, Pyysalo, Nagel, Bose, Muhammad, Sharma,
  Longpre, maieh Nikpoor, Silberberg, Pai, Zink, Torrent, Schick, Thrush,
  Danchev, Nikoulina, Laippala, Lepercq, Prabhu, Alyafeai, Talat, Raja,
  Heinzerling, Si, Salesky, Mielke, Lee, Sharma, Santilli, Chaffin, Stiegler,
  Datta, Szczechla, Chhablani, Wang, Pandey, Strobelt, Fries, Rozen, Gao,
  Sutawika, Bari, Al-Shaibani, Manica, Nayak, Teehan, Albanie, Shen, Ben-David,
  Bach, Kim, Bers, F{\'e}vry, Neeraj, Thakker, Raunak, Tang, Yong, Sun, Brody,
  Uri, Tojarieh, Roberts, Chung, Tae, Phang, Press, Li, Narayanan, Bourfoune,
  Casper, Rasley, Ryabinin, Mishra, Zhang, Shoeybi, Peyrounette, Patry, Tazi,
  Sanseviero, von Platen, Cornette, Lavall'ee, Lacroix, Rajbhandari, Gandhi,
  Smith, Requena, Patil, Dettmers, Baruwa, Singh, Cheveleva, Ligozat,
  Subramonian, N'ev'eol, Lovering, Garrette, Tunuguntla, Reiter, Taktasheva,
  Voloshina, Bogdanov, Winata, Schoelkopf, Kalo, Novikova, Forde, Tang, Kasai,
  Kawamura, Hazan, Carpuat, Clinciu, Kim, Cheng, Serikov, Antverg, van~der Wal,
  Zhang, Zhang, Gehrmann, Mirkin, Pais, Shavrina, Scialom, Yun, Limisiewicz,
  Rieser, Protasov, Mikhailov, Pruksachatkun, Belinkov, Bamberger, Kasner,
  Kasner, Pestana, Feizpour, Khan, Faranak, Santos, Hevia, Unldreaj, Aghagol,
  Abdollahi, Tammour, HajiHosseini, Behroozi, Ajibade, Saxena, Ferrandis,
  Contractor, Lansky, David, Kiela, Nguyen, Tan, Baylor, Ozoani, Mirza,
  Ononiwu, Rezanejad, Jones, Bhattacharya, Solaiman, Sedenko, Nejadgholi,
  Passmore, Seltzer, Sanz, Fort, Dutra, Samagaio, Elbadri, Mieskes, Gerchick,
  Akinlolu, McKenna, Qiu, Ghauri, Burynok, Abrar, Rajani, Elkott, Fahmy,
  Samuel, An, Kromann, Hao, Alizadeh, Shubber, Wang, Roy, Viguier, Le, Oyebade,
  Le, Yang, Nguyen, Kashyap, Palasciano, Callahan, Shukla, Miranda-Escalada,
  Singh, Beilharz, Wang, de~Brito, Zhou, Jain, Xu, Fourrier, Perin'an, Molano,
  Yu, Manjavacas, Barth, Fuhrimann, Altay, Bayrak, Burns, Vrabec, Bello, Dash,
  Kang, Giorgi, Golde, Posada, Sivaraman, Bulchandani, Liu, Shinzato,
  de~Bykhovetz, Takeuchi, P{\`a}mies, Castillo, Nezhurina, Sanger, Samwald,
  Cullan, Weinberg, Wolf, Mihaljcic, Liu, Freidank, Kang, Seelam, Dahlberg,
  Broad, Muellner, Fung, Haller, Chandrasekhar, Eisenberg, Martin, Canalli, Su,
  Su, Cahyawijaya, Garda, Deshmukh, Mishra, Kiblawi, Ott, Sang-aroonsiri,
  Kumar, Schweter, Bharati, Laud, Gigant, Kainuma, Kusa, Labrak, Bajaj,
  Venkatraman, Xu, Xu, Xu, Tan, Xie, Ye, Bras, Belkada, and Wolf]{bloom}
Scao, T.~L., Fan, A., Akiki, C., Pavlick, E., Ili'c, S., Hesslow, D.,
  Castagn'e, R., Luccioni, A.~S., Yvon, F., Gall{\'e}, M., Tow, J., Rush,
  A.~M., Biderman, S.~R., Webson, A., Ammanamanchi, P.~S., Wang, T., Sagot, B.,
  Muennighoff, N., del Moral, A.~V., Ruwase, O., Bawden, R., Bekman, S.,
  McMillan-Major, A., Beltagy, I., Nguyen, H., Saulnier, L., Tan, S., Suarez,
  P.~O., Sanh, V., Laurenccon, H., Jernite, Y., Launay, J., Mitchell, M.,
  Raffel, C., Gokaslan, A., Simhi, A., Etxabe, A.~S., Aji, A.~F., Alfassy, A.,
  Rogers, A., Nitzav, A.~K., Xu, C., Mou, C., Emezue, C.~C., Klamm, C., Leong,
  C., van Strien, D.~A., Adelani, D.~I., Radev, D.~R., Ponferrada, E.~G.,
  Levkovizh, E., Kim, E., Natan, E., Toni, F.~D., Dupont, G., Kruszewski, G.,
  Pistilli, G., ElSahar, H., Benyamina, H., Tran, H.~T., Yu, I., Abdulmumin,
  I., Johnson, I., Gonzalez-Dios, I., de~la Rosa, J., Chim, J., Dodge, J., Zhu,
  J., Chang, J., Frohberg, J., Tobing, J.~L., Bhattacharjee, J., Almubarak, K.,
  Chen, K., Lo, K., von Werra, L., Weber, L., Phan, L., Allal, L.~B., Tanguy,
  L., Dey, M., Mu{\~n}oz, M.~R., Masoud, M., Grandury, M., vSavsko, M., Huang,
  M., Coavoux, M., Singh, M., Jiang, M. T.-J., Vu, M.~C., Jauhar, M.~A.,
  Ghaleb, M., Subramani, N., Kassner, N., Khamis, N., Nguyen, O., Espejel, O.,
  de~Gibert, O., Villegas, P., Henderson, P., Colombo, P., Amuok, P., Lhoest,
  Q., Harliman, R., Bommasani, R., L'opez, R., Ribeiro, R., Osei, S., Pyysalo,
  S., Nagel, S., Bose, S., Muhammad, S.~H., Sharma, S.~S., Longpre, S., maieh
  Nikpoor, S., Silberberg, S., Pai, S., Zink, S., Torrent, T.~T., Schick, T.,
  Thrush, T., Danchev, V., Nikoulina, V., Laippala, V., Lepercq, V., Prabhu,
  V., Alyafeai, Z., Talat, Z., Raja, A., Heinzerling, B., Si, C., Salesky, E.,
  Mielke, S.~J., Lee, W.~Y., Sharma, A., Santilli, A., Chaffin, A., Stiegler,
  A., Datta, D., Szczechla, E., Chhablani, G., Wang, H., Pandey, H., Strobelt,
  H., Fries, J.~A., Rozen, J., Gao, L., Sutawika, L., Bari, M.~S., Al-Shaibani,
  M.~S., Manica, M., Nayak, N.~V., Teehan, R., Albanie, S., Shen, S.,
  Ben-David, S., Bach, S.~H., Kim, T., Bers, T., F{\'e}vry, T., Neeraj, T.,
  Thakker, U., Raunak, V., Tang, X., Yong, Z.-X., Sun, Z., Brody, S., Uri, Y.,
  Tojarieh, H., Roberts, A., Chung, H.~W., Tae, J., Phang, J., Press, O., Li,
  C., Narayanan, D., Bourfoune, H., Casper, J., Rasley, J., Ryabinin, M.,
  Mishra, M., Zhang, M., Shoeybi, M., Peyrounette, M., Patry, N., Tazi, N.,
  Sanseviero, O., von Platen, P., Cornette, P., Lavall'ee, P.~F., Lacroix, R.,
  Rajbhandari, S., Gandhi, S., Smith, S., Requena, S., Patil, S., Dettmers, T.,
  Baruwa, A., Singh, A., Cheveleva, A., Ligozat, A.-L., Subramonian, A.,
  N'ev'eol, A., Lovering, C., Garrette, D.~H., Tunuguntla, D.~R., Reiter, E.,
  Taktasheva, E., Voloshina, E., Bogdanov, E., Winata, G.~I., Schoelkopf, H.,
  Kalo, J.-C., Novikova, J., Forde, J.~Z., Tang, X., Kasai, J., Kawamura, K.,
  Hazan, L., Carpuat, M., Clinciu, M., Kim, N., Cheng, N., Serikov, O.,
  Antverg, O., van~der Wal, O., Zhang, R., Zhang, R., Gehrmann, S., Mirkin, S.,
  Pais, S.~O., Shavrina, T., Scialom, T., Yun, T., Limisiewicz, T., Rieser, V.,
  Protasov, V., Mikhailov, V., Pruksachatkun, Y., Belinkov, Y., Bamberger, Z.,
  Kasner, Z., Kasner, Z., Pestana, A., Feizpour, A., Khan, A., Faranak, A.,
  Santos, A. S.~R., Hevia, A., Unldreaj, A., Aghagol, A., Abdollahi, A.,
  Tammour, A., HajiHosseini, A., Behroozi, B., Ajibade, B.~A., Saxena, B.~K.,
  Ferrandis, C.~M., Contractor, D., Lansky, D.~M., David, D., Kiela, D.,
  Nguyen, D.~A., Tan, E., Baylor, E., Ozoani, E., Mirza, F.~T., Ononiwu, F.,
  Rezanejad, H., Jones, H., Bhattacharya, I., Solaiman, I., Sedenko, I.,
  Nejadgholi, I., Passmore, J., Seltzer, J., Sanz, J.~B., Fort, K., Dutra, L.,
  Samagaio, M., Elbadri, M., Mieskes, M., Gerchick, M., Akinlolu, M., McKenna,
  M., Qiu, M., Ghauri, M., Burynok, M., Abrar, N., Rajani, N., Elkott, N.,
  Fahmy, N., Samuel, O., An, R., Kromann, R.~P., Hao, R., Alizadeh, S.,
  Shubber, S., Wang, S.~L., Roy, S., Viguier, S., Le, T.-C., Oyebade, T., Le,
  T. N.~H., Yang, Y., Nguyen, Z.~K., Kashyap, A.~R., Palasciano, A., Callahan,
  A., Shukla, A., Miranda-Escalada, A., Singh, A.~K., Beilharz, B., Wang, B.,
  de~Brito, C. M.~F., Zhou, C., Jain, C., Xu, C., Fourrier, C., Perin'an,
  D.~L., Molano, D., Yu, D., Manjavacas, E., Barth, F., Fuhrimann, F., Altay,
  G., Bayrak, G., Burns, G., Vrabec, H.~U., Bello, I.~I., Dash, I., Kang,
  J.~S., Giorgi, J., Golde, J., Posada, J.~D., Sivaraman, K., Bulchandani, L.,
  Liu, L., Shinzato, L., de~Bykhovetz, M.~H., Takeuchi, M., P{\`a}mies, M.,
  Castillo, M.~A., Nezhurina, M., Sanger, M., Samwald, M., Cullan, M.,
  Weinberg, M., Wolf, M., Mihaljcic, M., Liu, M., Freidank, M., Kang, M.,
  Seelam, N., Dahlberg, N., Broad, N.~M., Muellner, N., Fung, P., Haller, P.,
  Chandrasekhar, R., Eisenberg, R., Martin, R., Canalli, R.~L., Su, R., Su, R.,
  Cahyawijaya, S., Garda, S., Deshmukh, S.~S., Mishra, S., Kiblawi, S., Ott,
  S., Sang-aroonsiri, S., Kumar, S., Schweter, S., Bharati, S.~P., Laud, T.,
  Gigant, T., Kainuma, T., Kusa, W., Labrak, Y., Bajaj, Y., Venkatraman, Y.,
  Xu, Y., Xu, Y., Xu, Y., Tan, Z.~X., Xie, Z., Ye, Z., Bras, M., Belkada, Y.,
  and Wolf, T.
\newblock Bloom: A 176b-parameter open-access multilingual language model.
\newblock \emph{ArXiv}, abs/2211.05100, 2022.
\newblock URL \url{https://api.semanticscholar.org/CorpusID:253420279}.

\bibitem[Sennrich et~al.(2016)Sennrich, Haddow, and
  Birch]{sennrich-etal-2016-neural}
Sennrich, R., Haddow, B., and Birch, A.
\newblock Neural machine translation of rare words with subword units.
\newblock In \emph{Proceedings of the 54th Annual Meeting of the Association
  for Computational Linguistics (Volume 1: Long Papers)}, pp.\  1715--1725,
  Berlin, Germany, August 2016. Association for Computational Linguistics.
\newblock \doi{10.18653/v1/P16-1162}.
\newblock URL \url{https://aclanthology.org/P16-1162}.

\bibitem[Thawani et~al.(2021)Thawani, Pujara, Ilievski, and
  Szekely]{thawani-etal-2021-representing}
Thawani, A., Pujara, J., Ilievski, F., and Szekely, P.
\newblock Representing numbers in {NLP}: a survey and a vision.
\newblock In \emph{Proceedings of the 2021 Conference of the North American
  Chapter of the Association for Computational Linguistics: Human Language
  Technologies}, pp.\  644--656, Online, June 2021. Association for
  Computational Linguistics.
\newblock \doi{10.18653/v1/2021.naacl-main.53}.
\newblock URL \url{https://aclanthology.org/2021.naacl-main.53}.

\bibitem[Touvron et~al.(2023{\natexlab{a}})Touvron, Lavril, Izacard, Martinet,
  Lachaux, Lacroix, Rozi{\`{e}}re, Goyal, Hambro, Azhar, Rodriguez, Joulin,
  Grave, and Lample]{touvron2023llama}
Touvron, H., Lavril, T., Izacard, G., Martinet, X., Lachaux, M., Lacroix, T.,
  Rozi{\`{e}}re, B., Goyal, N., Hambro, E., Azhar, F., Rodriguez, A., Joulin,
  A., Grave, E., and Lample, G.
\newblock Llama: Open and efficient foundation language models.
\newblock \emph{CoRR}, abs/2302.13971, 2023{\natexlab{a}}.
\newblock \doi{10.48550/ARXIV.2302.13971}.
\newblock URL \url{https://doi.org/10.48550/arXiv.2302.13971}.

\bibitem[Touvron et~al.(2023{\natexlab{b}})Touvron, Martin, Stone, Albert,
  Almahairi, Babaei, Bashlykov, Batra, Bhargava, Bhosale, Bikel, Blecher,
  Canton{-}Ferrer, Chen, Cucurull, Esiobu, Fernandes, Fu, Fu, Fuller, Gao,
  Goswami, Goyal, Hartshorn, Hosseini, Hou, Inan, Kardas, Kerkez, Khabsa,
  Kloumann, Korenev, Koura, Lachaux, Lavril, Lee, Liskovich, Lu, Mao, Martinet,
  Mihaylov, Mishra, Molybog, Nie, Poulton, Reizenstein, Rungta, Saladi,
  Schelten, Silva, Smith, Subramanian, Tan, Tang, Taylor, Williams, Kuan, Xu,
  Yan, Zarov, Zhang, Fan, Kambadur, Narang, Rodriguez, Stojnic, Edunov, and
  Scialom]{touvron2023llama2}
Touvron, H., Martin, L., Stone, K., Albert, P., Almahairi, A., Babaei, Y.,
  Bashlykov, N., Batra, S., Bhargava, P., Bhosale, S., Bikel, D., Blecher, L.,
  Canton{-}Ferrer, C., Chen, M., Cucurull, G., Esiobu, D., Fernandes, J., Fu,
  J., Fu, W., Fuller, B., Gao, C., Goswami, V., Goyal, N., Hartshorn, A.,
  Hosseini, S., Hou, R., Inan, H., Kardas, M., Kerkez, V., Khabsa, M.,
  Kloumann, I., Korenev, A., Koura, P.~S., Lachaux, M., Lavril, T., Lee, J.,
  Liskovich, D., Lu, Y., Mao, Y., Martinet, X., Mihaylov, T., Mishra, P.,
  Molybog, I., Nie, Y., Poulton, A., Reizenstein, J., Rungta, R., Saladi, K.,
  Schelten, A., Silva, R., Smith, E.~M., Subramanian, R., Tan, X.~E., Tang, B.,
  Taylor, R., Williams, A., Kuan, J.~X., Xu, P., Yan, Z., Zarov, I., Zhang, Y.,
  Fan, A., Kambadur, M., Narang, S., Rodriguez, A., Stojnic, R., Edunov, S.,
  and Scialom, T.
\newblock Llama 2: Open foundation and fine-tuned chat models.
\newblock \emph{CoRR}, abs/2307.09288, 2023{\natexlab{b}}.
\newblock \doi{10.48550/ARXIV.2307.09288}.
\newblock URL \url{https://doi.org/10.48550/arXiv.2307.09288}.

\bibitem[Wang et~al.(2021)Wang, Wang, Joty, and Hoi]{Wang_Wang_Joty_Hoi_2021}
Wang, Y., Wang, W., Joty, S., and Hoi, S.~C.
\newblock Codet5: Identifier-aware unified pre-trained encoder-decoder models
  for code understanding and generation.
\newblock In \emph{Proceedings of the 2021 Conference on Empirical Methods in
  Natural Language Processing}, pp.\  8696–8708, Online and Punta Cana,
  Dominican Republic, Nov 2021. Association for Computational Linguistics.
\newblock \doi{10.18653/v1/2021.emnlp-main.685}.
\newblock URL \url{https://aclanthology.org/2021.emnlp-main.685}.

\bibitem[Wenzek et~al.(2020)Wenzek, Lachaux, Conneau, Chaudhary, Guzm{\'a}n,
  Joulin, and Grave]{ccnet}
Wenzek, G., Lachaux, M.-A., Conneau, A., Chaudhary, V., Guzm{\'a}n, F., Joulin,
  A., and Grave, E.
\newblock {CCN}et: Extracting high quality monolingual datasets from web crawl
  data.
\newblock In Calzolari, N., B{\'e}chet, F., Blache, P., Choukri, K., Cieri, C.,
  Declerck, T., Goggi, S., Isahara, H., Maegaard, B., Mariani, J., Mazo, H.,
  Moreno, A., Odijk, J., and Piperidis, S. (eds.), \emph{Proceedings of the
  Twelfth Language Resources and Evaluation Conference}, pp.\  4003--4012,
  Marseille, France, May 2020. European Language Resources Association.
\newblock ISBN 979-10-95546-34-4.
\newblock URL \url{https://aclanthology.org/2020.lrec-1.494}.

\bibitem[Wolf et~al.(2019)Wolf, Debut, Sanh, Chaumond, Delangue, Moi, Cistac,
  Rault, Louf, Funtowicz, and Brew]{Wolf2019HuggingFacesTS}
Wolf, T., Debut, L., Sanh, V., Chaumond, J., Delangue, C., Moi, A., Cistac, P.,
  Rault, T., Louf, R., Funtowicz, M., and Brew, J.
\newblock Huggingface's transformers: State-of-the-art natural language
  processing.
\newblock \emph{CoRR}, abs/1910.03771, 2019.
\newblock URL \url{http://arxiv.org/abs/1910.03771}.

\bibitem[Xue et~al.(2022)Xue, Barua, Constant, Al-Rfou, Narang, Kale, Roberts,
  and Raffel]{xue-etal-2022-byt5}
Xue, L., Barua, A., Constant, N., Al-Rfou, R., Narang, S., Kale, M., Roberts,
  A., and Raffel, C.
\newblock {B}y{T}5: Towards a token-free future with pre-trained byte-to-byte
  models.
\newblock \emph{Transactions of the Association for Computational Linguistics},
  10:\penalty0 291--306, 2022.
\newblock \doi{10.1162/tacl_a_00461}.
\newblock URL \url{https://aclanthology.org/2022.tacl-1.17}.

\bibitem[Zheng et~al.(2023)Zheng, Xia, Zou, Dong, Wang, Xue, Shen, Wang, Wang,
  Li, Su, Yang, and Tang]{CodeGeeX}
Zheng, Q., Xia, X., Zou, X., Dong, Y., Wang, S., Xue, Y., Shen, L., Wang, Z.,
  Wang, A., Li, Y., Su, T., Yang, Z., and Tang, J.
\newblock Codegeex: {A} pre-trained model for code generation with multilingual
  benchmarking on humaneval-x.
\newblock In Singh, A.~K., Sun, Y., Akoglu, L., Gunopulos, D., Yan, X., Kumar,
  R., Ozcan, F., and Ye, J. (eds.), \emph{Proceedings of the 29th {ACM}
  {SIGKDD} Conference on Knowledge Discovery and Data Mining, {KDD} 2023, Long
  Beach, CA, USA, August 6-10, 2023}, pp.\  5673--5684. {ACM}, 2023.
\newblock \doi{10.1145/3580305.3599790}.
\newblock URL \url{https://doi.org/10.1145/3580305.3599790}.

\bibitem[Zipf(1949)]{zipf1949human}
Zipf, G.~K.
\newblock \emph{Human Behavior and the Principle of Least Effort}.
\newblock Addison-Wesley, 1949.

\bibitem[Zouhar et~al.(2023)Zouhar, Meister, Gastaldi, Du, Sachan, and
  Cotterell]{Zouhar_Meister_Gastaldi_Du_Sachan_Cotterell_2023}
Zouhar, V., Meister, C., Gastaldi, J.~L., Du, L., Sachan, M., and Cotterell, R.
\newblock Tokenization and the noiseless channel.
\newblock In Rogers, A., Boyd{-}Graber, J.~L., and Okazaki, N. (eds.),
  \emph{Proceedings of the 61st Annual Meeting of the Association for
  Computational Linguistics (Volume 1: Long Papers), {ACL} 2023, Toronto,
  Canada, July 9-14, 2023}, pp.\  5184--5207. Association for Computational
  Linguistics, 2023.
\newblock \doi{10.18653/V1/2023.ACL-LONG.284}.
\newblock URL \url{https://doi.org/10.18653/v1/2023.acl-long.284}.

\end{thebibliography}
\bibliographystyle{icml2024}

\newpage
\appendix

\begin{figure}
    \centering
    \includegraphics[width=\linewidth]{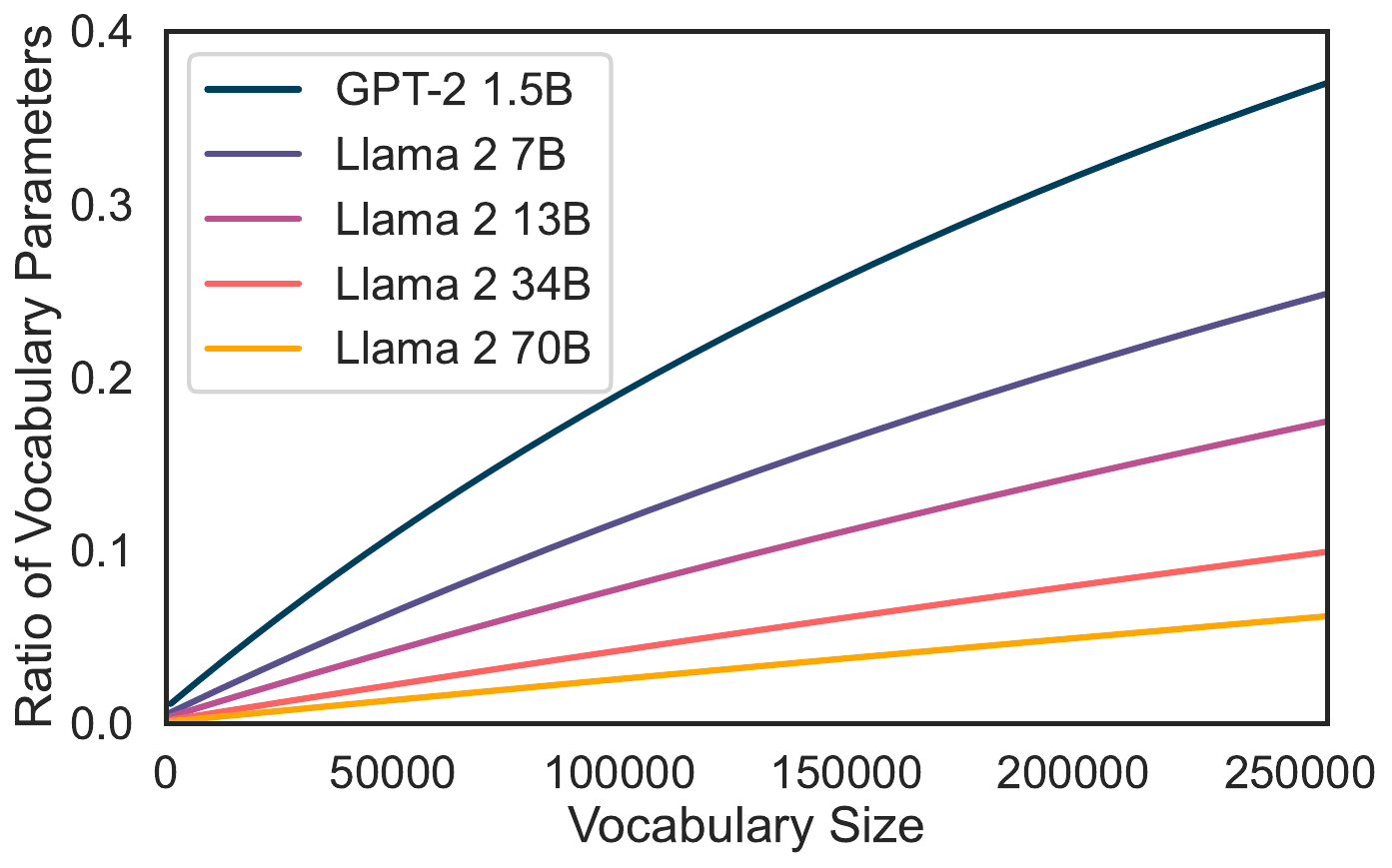}
    \caption{Proportion of parameters used by different vocabulary sizes for different model sizes of \llama~2 and GPT-2. 
    }
    \label{fig:memory_vocab}
\end{figure}

\section{Compression Evaluation Dataset}
\label{sec:data}

Our English held-out dataset is 1000 examples from CCnet \cite{ccnet} and 1000 examples from the English subset of Wikipedia.

To create our multilingual evaluation set, we hold-out 1000 examples for each language present in the Wikipedia dataset. 
This constitutes of 1000 articles from the following languages: Italian, Bulgarian, Egyptian Arabic, Spanish, Serbian, German, Ukrainian, Arabic, Czech, Catalan, Slovenian, Persian, Polish, Russian, Finnish, Swedish, Portuguese, French, Japanese, Croatian, Chinese, Romanian, Dutch, Indonesian, Hungarian, Korean, Danish, and Vietnamese.

To create our code evaluation dataset, we hold-out 1000 files from the 30 most popular programming languages in the Stack dataset \cite{Kocetkov2022TheS3}: Assembly, Batchfile, C, C\#, C++, CMake, CSS, Dockerfile, FORTRAN, GO, HTML, Haskell, Java, JavaScript, Julia, Lua, Makefile, Markdown, PHP, Perl, PowerShell, Python, Ruby, Rust, SQL, Scala, Shell, TeX, TypeScript, and Visual Basic.

\section{Vocabulary Size and Model Size}
\label{sec:vocab-size}

We show the proportion of parameters used by tokenizers of different vocabulary sizes for different model sizes of \llama~2 and GPT-2 for scale in Figure~\ref{fig:memory_vocab}.
This shows that expanding the number of parameters in a LLM of size 70B only marginally increases the total size of the model and the ratio of vocab parameters to model parameters.
While increasing the vocabulary size for smaller models, such as a GPT-2 XL 1.5B model, has a much larger effect on the total number of parameters.

\section{Tokenizer Re-use}
\label{sec:tokenizer_reuse}

\begin{figure*}
    \centering
    \includegraphics[width=\linewidth]{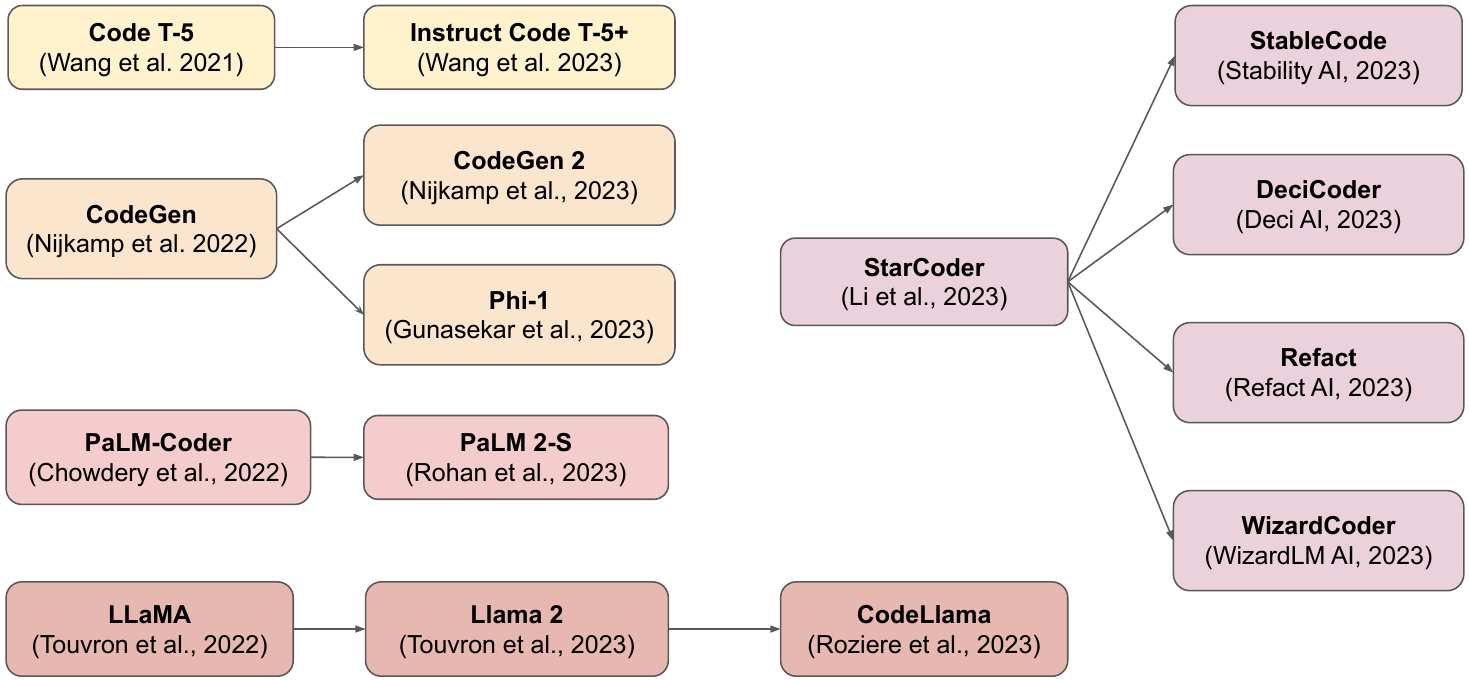}
    \caption{\textbf{Tokenizer re-use is common across code generation LLMs.}}
    \label{fig:tokenizer_reuse}
\end{figure*}

Figure~\ref{fig:tokenizer_reuse} shows examples of tokenizer re-use across multiple SOTA code generation models.
This is not a comprehensive mapping, but serves to illustrate how often tokenization is left as an implementation detail that is rarely innovated upon.

\section{Normalization}
\label{sec:normalization}

Normalization is a pre-tokenization step that aims to reduce the complexity of text data by converting it into a unified format. 
Different forms of UTF-8 normalizations have both been used as defaults by different tokenizer libraries -- Sentencepiece \cite{Kudo2018SentencePieceAS} uses Normalization Form KC (NFKC) and TokenMonster \cite{tokenmonster} uses Normalization Form D (NFD).

\begin{itemize}
    \item \textbf{NFKC Normalization}: In NFKC, characters are transformed to their most common, compatible form. 
    For instance, a superscript UTF-8 number will get converted to a standard number: $\text{NFKC}(\empty^2) = 2$
    \item \textbf{NFD Normalization}: Conversely, NFD decomposes characters into their basic components, separating letters from their diacritical marks. 
    For example, a character with an accent would be split into its letter and accent components: $\text{NFD}(\tilde{n}) = n + \tilde{ }$
\end{itemize}

Despite the potential advantages of a normalization step, these are generally not reversible transformations\footnote{NFD is mostly reversible if the input text is in form C but has rare edge cases.}. %
Therefore the tokens processed by the tokenizer may not retain their original form, violating $x = \text{encode}(\text{decode}(x))$.
This can lead to sub-optimal normalization, where for instance converting the superscript $\empty^2$ to its base representation $2$ throws away the power semantic information -- which could be crucial for mathematical reasoning.
We forgo the use of any normalization methods in our proposed tokenizers and broadly recommend that modern tokenizers be reversible.

\section{The \punct Tokenizer}
\label{sec:punct}

In \punct, we remove the \gpt English-specific contractions and prevent certain white-spaces and punctuation tokens such as \texttt{\textbackslash t} or \texttt{.} to be encoded at the start of a letter-only token.
The \punct pre-tokenizer regular expression is shown in Figure~\ref{fig:regexex}, and, compared to \gpt trades away compression in exchange for composition.
For instance, \punct prevents any form of tokens that starts with a \texttt{.} (period) which is a common shorthand to accessing attributes in many programming languages.
Consider the \texttt{.append} method in python, the \gpt tokenizer would tokenize the entire sequence \texttt{.append} as a single token while \punct would tokenize \texttt{.} and \texttt{append} separately.
We use \punct to test whether a stronger separation between syntax and semantics, simplifies the language generation task and translates to greater downstream performance.

In our initial results, shown in Table~\ref{tab:finetuning-32k-all}, we found \punct\textsuperscript{NL} 32k to be better than both \gpt\textsuperscript{NL} 32k and \llama\textsuperscript{NL} on Pass@1 performance in both HumanEval and MBPP.
However, the results obtained on a 7B model (Table~\ref{tab:7b}) using \punct 80k show no significant edge over the \codellama baseline and either the \merged or \gpt 80k tokenizers.
We thus cannot conclude that a stronger separation between syntax and semantics leads to greater downstream performance. 
Instead, we would recommend usage of the \gpt regular expression tokenizer since it provides an additional $5\%$ compression over \punct at no significant cost to performance.

\section{Tokenizer Compression of Popular LLMs}

Table~\ref{tab:code-compression-all} shows the tokenizer compression rates for a large number of popular LLMs.
Most, but not all, of the models selected are models that have been promoted or advertised for code generation.
We also include our own tokenizers for comparison.

\section{Examples of tokenization}
\label{sec:example}

We compare different tokenizers on a sample code in Table~\ref{tab:encodings}. 
In our example snippet, we can see the effects of the different pre-tokenization schemes in practice.
For instance, GPT-2 did not restrict the max number of consecutive digits in its pre-tokenization and therefore has a token to represent `1000', whereas the \gpt and \punct regular expression limit the maximum number of digits to 3, and therefore encode `1000' as `100' and `0'.
\llama uses single digits and thus uses 4 tokens to encode the number `1000'.
We also see that the \gpt pre-tokenization allows learning tokens such as `.append', which \punct cannot.

We can also observe the effects of having an unrestricted pre-tokenization scheme such as \identity, which even learn a token for the numpy import statement including the new line symbol. 
These types of savings allows \identity to save more than $50\%$ tokens compared to \llama. 
However, we can also observe the downsides of the \identity tokenizer in its first tokenization of `tqdm~'.
Since the BPE merge rule for `m' and `~' comes first, `tqdm~' is tokenized as `tq',`d' and `m~' separately.
Whereas the second instance of `tqdm', which is followed by a new line token, is able to be tokenized into a single token `tqdm'.

\section{Training \smallmodel and \smallcodemodel~1.5B Base Models}
\label{sec:base-models} 

To test our hypotheses, we train two GPT-2 XL 1.5B base models \cite{gpt2} which we refer to as \smallmodel~1.5B and \smallcodemodel~1.5B.
\smallmodel~1.5B is trained for 2T tokens on a general data-mix resembling that of \llama~2, while \smallcodemodel~1.5B is fine-tuned from \smallmodel~1.5B for 500B tokens using a code-specific data-mix.
We use the same hyper-parameters as in \llama~2~\cite{touvron2023llama2} and \codellama~\cite{roziere2023code}, see Table~\ref{tab:gpt2-xl-config} for an overview.

\section{Rényi Entropy in Code Tokenizers}
\label{sec:renyi}

As shown in Table~\ref{tab:code-compression-all}, following the methodology of \citet{Zouhar_Meister_Gastaldi_Du_Sachan_Cotterell_2023}, we measure the Rényi entropy of all tokenizers with an $\alpha = 2.5$. 
Rényi entropy quantifies the evenness in the distribution of tokens, penalizing those distributions with highly uneven token frequencies. \citet{Zouhar_Meister_Gastaldi_Du_Sachan_Cotterell_2023} reported a correlation between Rényi entropy and downstream performance, as indicated by the BLEU score in a Machine Translation task.

However, our analysis with trained 1.5B models reveals a different trend. 
We found a significant negative correlation between Rényi entropy and Human Eval Pass@1 (Pearson $r=-0.7765, p=0.040$) and MBPP Pass@1 (Pearson $r=-0.9007, p=0.0057$). 
This suggests that a higher Rényi entropy ($\alpha = 2.5$), contrary to expectations, correlates with lower performance in these code generation metrics.
Notably, the \identity tokenizer, which performed worst in our evaluations, showed the highest Rényi entropy. 
This can be attributed to the lack of a pre-tokenizer, allowing the BPE algorithm to more evenly distribute tokens according to frequency.

\section{Word-Equivalent Results}
\label{sec:word-equivalent}

We report our results for both \textit{token-equivalent} and \textit{word-equivalent} metrics in Table~\ref{tab:finetuning-32k-all}.
Word-equivalent compares the models after being trained on the same amount of information as measured with the number of characters.
Due to added compression of our tokenizers compared to \llama, the word-equivalent evaluation corresponds to training on less tokens and using less compute than the token-equivalent evaluation.

\section{Freezing Weights}
\label{freezing_weights}

\begin{table}[h]
\begin{tabular}{lrrrr}
\toprule
& \multicolumn{2}{c}{\textbf{Human Eval}} & \multicolumn{2}{c}{\textbf{MBPP}} \\
 & \multicolumn{1}{l}{\textbf{Pass@1}} & \multicolumn{1}{l}{\textbf{Pass@100}} & \multicolumn{1}{l}{\textbf{Pass@1}} & \multicolumn{1}{l}{\textbf{Pass@100}} \\
\midrule
0B & \acc{22.22560976} & \acc{68.7193931} & \acc{29.75} & \acc{73.5575} \\
100B & \acc{21.33231707} & \acc{67.45936547} & \acc{28.527} & \acc{72.32964063} \\
500B & \acc{17.80792683} & \acc{59.66465796} & \acc{24.482} & \acc{68.88526563} \\
\bottomrule
\end{tabular}
\caption{
\textbf{Keeping LLM weights frozen for a number of tokens.} 
We compare the performance of \punct\textsuperscript{NL} with weights frozen for different numbers of tokens and report results are on Pass@1 and Pass@100 for Human Evaluation and MBPP. 
The rows (0B, 100B, 500B) represent the number of tokens for which weights were frozen for.
We find freezing weights ineffective to adapt a LLM to a change in its tokenizer.
}
\label{tab:freezing}
\end{table}

Since the baseline model is assumed to be a pre-trained LLM, we also experiment with freezing the weights of the model that do not pertain to the vocabulary and training only the weights affected by the change -- the embedding and output layers.
We test two different setups where we 1) freeze the models weights except the embedding/output layers only for 100B tokens and then unfreeze the entire model for the remaining 400B tokens, and 2)freeze the model weights except the embedding/output layers for the 500B tokens.
Our motivation for 1) is to test whether the model benefits from an initial alignment of its input/outputs to its intermediate representations, and for 2) is to test whether the model can learn to adapt its weights to the new input/outputs it receives without changing its internal structure.

Table~\ref{tab:freezing} shows that only training only training the embedding and output weights for 100B (20\% of training) or for 500B (100\% of training) leads to worse performance on all downstream metrics. 
We find freezing weights to be an ineffective strategy to aligning the vocabulary-specific weights of the model with its internal structure.

\begin{table*}[h]
\centering
\begin{tabular}{l|c|c}
\hline
\textbf{Configuration} & \textbf{\smallmodel~1.5B} & \textbf{\smallcodemodel~1.5B}\\
\hline
Initialization & Random & \smallmodel~1.5B \\
Number of Parameters & 1.5B & 1.5B\\
Number of Layers & 48 & 48\\
Hidden Size & 1600 & 1600\\
Number of Attention Heads & 25 & 25\\
Sequence Length & 4096 & 4096 \\
Optimizer & AdamW \cite{Loshchilov2017DecoupledWD} & AdamW \cite{Loshchilov2017DecoupledWD} \\
Optimizer Parameters & $\beta_1 = 0.9, \beta_2 = 0.95$ & $\beta_1 = 0.9, \beta_2 = 0.95$ \\
Learning Rate & $3 \times 10^{-4}$ & $3 \times 10^{-4}$ \\
Learning Rate Schedule & Cosine & Cosine  \\
Learning Rate Decay & $\frac{1}{10}lr$ & $\frac{1}{30}lr$ \\
Number of Tokens Seen & 2T & 500B \\ 
Data & General & Code-specific \\
\hline
\end{tabular}
\caption{\smallmodel~1.5B Pre-Training configurations. Globally, we re-use the same parameters as \llama~2~\cite{touvron2023llama2} and \codellama~\cite{roziere2023code} and only change the architecture to the smaller 1.5B GPT-2 XL model type.}
\label{tab:gpt2-xl-config}
\end{table*}

\section{Tricks for more robust Tokenizers}
\label{sec:tricks}

\begin{figure}[H]
    \centering
    \scalebox{.3}{
    \includegraphics{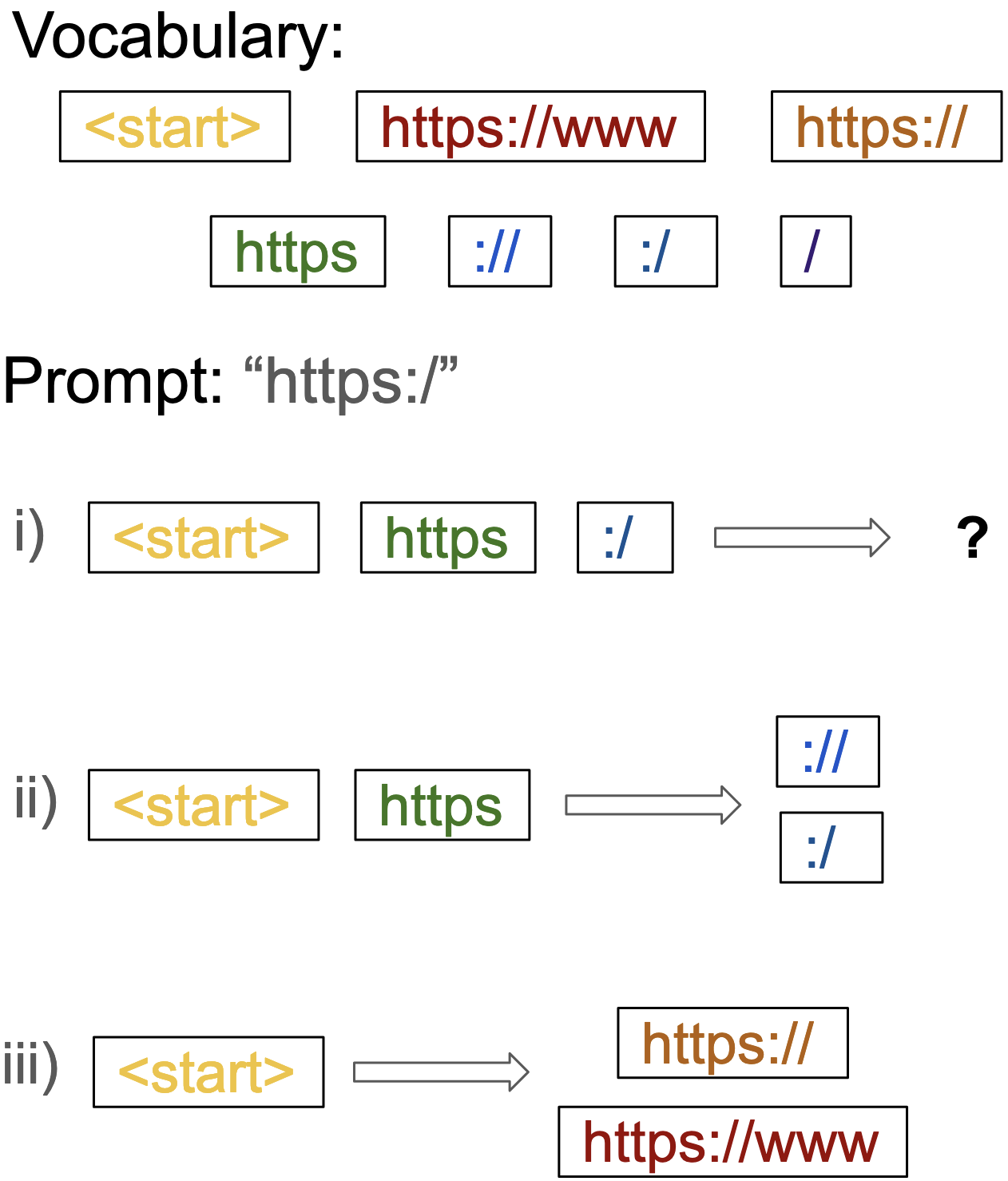}
    }
    \caption{\textbf{Token healing strategies.} We consider three decoding strategies given the prompt ``https:/''. \textbf{i)} is the decoding case without token healing, where the next decoding step is unconstrained. \textbf{ii)} is a single token healing step, where the prompt is backtracked by a single token and a single constrained decoding step is done over the possible tokens that fit the observed text (``:/''). \textbf{iii)} is the case with N backtracking steps, where the prompt is backtracked until there does not exist a token in the vocabulary which can cover the observation (``https:/''). A single constrained decoding step is then done over the possible tokens that fit the observation. The $N$ steps token healing method can therefore lead to different decoding outcomes than the single step backtracking.}
    \label{fig:token-healing}
\end{figure}

\subsection{Token healing}
\label{sec:token-healing}

\begin{table}[H]
\scalebox{0.85}{
\begin{tabular}{lrrrr}
\toprule
& \multicolumn{2}{c}{\textbf{Human Eval}} & \multicolumn{2}{c}{\textbf{MBPP}} \\
 & \multicolumn{1}{l}{\textbf{Pass@1}} & \multicolumn{1}{l}{\textbf{Pass@100}} & \multicolumn{1}{l}{\textbf{Pass@1}} & \multicolumn{1}{l}{\textbf{Pass@100}} \\
 \midrule
\multicolumn{5}{l}{Without token healing}\\
\llama\textsuperscript{C} & \acc{22.58231707} & \acc{67.60766602} & \acc{30.04} & \acc{74.07670313} \\
\gpt\textsuperscript{C} & \acc{21.4054878} & \acc{67.63908275} & \acc{28.772} & \acc{74.63954688} \\
\identitytable\textsuperscript{C} & \acc{0.1341463415} & \acc{3.509513669} & \acc{23.579} & \acc{68.78103125} \\
\midrule
\multicolumn{5}{l}{With token healing}\\
\llama\textsuperscript{C} & \acc{22.80182927} & \acc{68.60337867} & \acc{30.138} & \acc{74.351} \\
\gpt\textsuperscript{C} & \acc{21.2347561} & \acc{67.48544684} & \acc{28.886} & \acc{74.86586719} \\
\identitytable\textsuperscript{C} & \acc{18.77439024} & \acc{66.06609661} & \acc{23.409} & \acc{67.75559375} \\
\bottomrule
\end{tabular}
}
\caption{
\textbf{Token Healing Results.} We show results for three different tokenizers of size 32k trained from \codellama for 500B tokens. For both \llama and \gpt, token healing has very little impact on the evaluation metrics. Using the \identity tokenizer, the LLM is mostly unable to generate valid code HumanEval if not using token healing. This is likely because the prompt boundary in HumanEval splits a token learned in the \identity tokenizer, and thus disturbs the expected token distribution.
}
\label{tab:token-healing}
\end{table}

Token healing is a technique used to address biases introduced by tokenization at prompt boundaries~\cite{guidance_ai_guidance_2024}.
This bias occurs when the prompt ends with a string that could be encoded as the beginning of a token, leading to a skewed distribution. 
Token healing backs up the generation process by one token before the end of the prompt and constrains the first token generated to have a prefix that matches the last token in the prompt to mitigate this bias.
We find token healing to have a negligible effect on tokenizers with well-defined word/token boundaries, however token healing has a large impact on tokenizers such as the \identity tokenizer where alphanumeric tokens are allowed to include new lines and spaces (which are common separators in prompts).

See Figure~\ref{fig:token-healing} for an illustrated example of different token healing strategies.
We implement a version of token healing that steps back $N$ tokens until a valid token boundary is found.
This is different from the original implementation of token healing~\cite{guidance_ai_guidance_2024} that naively steps back by a single step.
This is because some of the tokenizers we test, such as the identity tokenizer, can contain tokens that are entire sentences, and simply doing a single step back might not allow that sentence token to be recovered.

We also quantitatively assess the impact of token healing by running our evaluations with and without token healing. 
Our results shown in Table~\ref{tab:token-healing} indicate that token healing is particularly effective for tokenizers with less defined boundaries, such as the \identity tokenizer.
In fact, the LLMs using the \identity tokenizer are unable to generate valid code given the prompt in HumanEval without token healing.

While token healing shows promise in addressing tokenization bias, it is not without limitations.
One key challenge is determining the optimal constrained decoding path for a given prompt.
As we show in Figure~\ref{fig:token-healing}, different numbers of backtracks might lead to different decoding outcomes.
In fact, to truly approximate the text distribution, a more correct implementation of token healing would have to implement a constrained beam-search.
One would have to backtrack by N steps and do multiple constrained decoding steps over the vocabulary to allow the smaller partial but potentially valid tokens to be generated.
We leave further exploration of token healing and beam-search implementation as future work.

\subsection{BPE dropout}

Another trick that one can use to adapt to new tokenizers or make a model more robust in its tokenization scheme is BPE dropout ~\cite{Provilkov_Emelianenko_Voita_2020}.
BPE dropout randomly drops merge entries in the BPE merge table, thus injecting noise in the tokenization process.
This leads to sentences being encoded sub-optimally, but with more variation in the token distribution than would naturally be observed with greedy encoding.
A model would therefore be able to see the same sentence being encoded in multiple different ways.

\citet{Provilkov_Emelianenko_Voita_2020} observe that BPE dropout improves MT performance, even in high-resource settings, and particularly when applied to noisy sources. 
Additionally, BPE dropout seems to produce more fine-grained segmentation, alleviate issues with frequent sequences of characters, and lead to better embedding spaces compared to traditional BPE.
Ultimately, BPE dropout makes the LLM more robust to specific tokenization and to prompt boundaries where the tokens encoded might not reflect the observed token distribution.

\begin{table*}[h]
\centering
\scalebox{0.9}{
\begin{tabular}{cc|cccc|cccc|c}
\toprule
\multicolumn{2}{r}{} & \multicolumn{4}{c}{Normalized Sequence Length (↓)} & \multicolumn{4}{c}{Bytes per Token (↑)} &  \\
 & \makecell{Vocabulary\\Size} & Avg. & Code & Eng. & Mult. & Avg. & Code & Eng. & Mult. & \makecell{Avg. Renyi \\ ($\alpha=2.5$)} \\
\midrule
\makecell{ByT5 \cite{xue-etal-2022-byt5}} & 256 & 3.10 & 2.70 & 3.82 & 2.79 & 1.00 & 1.00 & 1.00 & 1.00 & 0.51 \\
\makecell{CodeT5 \cite{Wang_Wang_Joty_Hoi_2021}} & 32k & 1.29 & 0.94 & 1.11 & 1.83 & 2.69 & 2.89 & 3.45 & 1.72 & 0.44 \\
\makecell{GPT-2 \cite{gpt2}} & 50k & 1.13 & 1.19 & 0.86 & 1.33 & 2.97 & 2.28 & 4.44 & 2.19 & 0.38 \\
\makecell{DeepSeek Coder \cite{deepseekai_2023_deepseekcoder13binstruct}} & 32k & 1.06 & 1.00 & 0.98 & 1.19 & 3.00 & 2.70 & 3.89 & 2.40 & 0.45 \\
\makecell{CodeGen \cite{Nijkamp_Pang_Hayashi_Tu_Wang_Zhou_Savarese_Xiong_2023}} & 50k & 1.05 & 0.95 & 0.86 & 1.33 & 3.16 & 2.83 & 4.44 & 2.19 & 0.43 \\
\makecell{SantaCoder \cite{santacoder}} & 49k & 1.04 & 0.88 & 1.07 & 1.17 & 3.01 & 3.08 & 3.59 & 2.36 & 0.47 \\
\makecell{Yi-6B \cite{01ai_2023_yi6b}} & 64k & 1.03 & 1.00 & 0.93 & 1.16 & 3.07 & 2.70 & 4.10 & 2.43 & 0.43 \\
\makecell{InCoder \cite{incoder}} & 50k & 1.03 & 0.74 & 1.02 & 1.31 & 3.17 & 3.66 & 3.73 & 2.11 & 0.51 \\
\makecell{Falcon \cite{falcon40b}} & 65k & 1.00 & 0.94 & 0.88 & 1.19 & 3.23 & 2.89 & 4.33 & 2.47 & 0.42 \\
\makecell{Persimmon \cite{persimmon-8b}} & 192k & 1.00 & 0.95 & 0.74 & 1.32 & 3.43 & 2.84 & 5.19 & 2.24 & 0.34 \\
\makecell{Replit \cite{replit}} & 32k & 1.00 & 0.85 & 1.06 & 1.10 & 3.10 & 3.19 & 3.60 & 2.51 & 0.48 \\
\makecell{\llama \cite{touvron2023llama}} & 32k & 1.00 & 1.00 & 1.00 & 1.00 & 3.10 & 2.70 & 3.82 & 2.79 & 0.46 \\
\makecell{Mistral \cite{jiang2023mistral}} & 32k & 0.99 & 0.99 & 0.97 & 1.02 & 3.13 & 2.74 & 3.93 & 2.72 & 0.45 \\
\makecell{Starcoder \cite{starcoder}} & 49k & 0.99 & 0.87 & 1.04 & 1.07 & 3.13 & 3.13 & 3.69 & 2.58 & 0.47 \\
\makecell{DeciCoder \cite{decicoder}} & 49k & 0.99 & 0.87 & 1.04 & 1.07 & 3.13 & 3.13 & 3.69 & 2.58 & 0.47 \\
\makecell{\puncttable } & 32k & 0.98 & 0.86 & 0.96 & 1.11 & 3.20 & 3.15 & 3.98 & 2.48 & 0.48 \\
\makecell{\gpt } & 32k & 0.97 & 0.81 & 0.97 & 1.13 & 3.24 & 3.35 & 3.95 & 2.44 & 0.50 \\
\makecell{gpt-neox \cite{Black2022GPTNeoX20BAO}} & 50k & 0.93 & 0.90 & 0.86 & 1.02 & 3.37 & 2.98 & 4.42 & 2.71 & 0.45 \\
\makecell{Pythia \cite{Biderman2023PythiaAS}} & 50k & 0.93 & 0.90 & 0.86 & 1.02 & 3.37 & 2.98 & 4.42 & 2.71 & 0.45 \\
\makecell{MPT \cite{mpt}} & 50k & 0.93 & 0.90 & 0.86 & 1.02 & 3.37 & 2.98 & 4.42 & 2.71 & 0.45 \\
\makecell{\identitytable} & 32k & 0.92 & 0.69 & 0.89 & 1.16 & 3.54 & 3.94 & 4.30 & 2.37 & 0.61 \\
\makecell{Claude \cite{anthropic_2023_claude}} & 65k & 0.91 & 0.84 & 0.87 & 1.04 & 3.43 & 3.22 & 4.41 & 2.66 & 0.44 \\
\makecell{\puncttable} & 64k & 0.90 & 0.82 & 0.89 & 0.99 & 3.45 & 3.29 & 4.28 & 2.79 & 0.46 \\
\makecell{\mergedtable} & 80k & 0.90 & 0.80 & 0.95 & 0.94 & 3.45 & 3.42 & 4.01 & 2.93 & 0.48 \\
\makecell{\gpt} & 64k & 0.89 & 0.76 & 0.90 & 1.01 & 3.52 & 3.55 & 4.27 & 2.72 & 0.49 \\
\makecell{\puncttable} & 80k & 0.88 & 0.81 & 0.88 & 0.95 & 3.52 & 3.32 & 4.35 & 2.89 & 0.46 \\
\makecell{\gpt} & 80k & 0.87 & 0.75 & 0.88 & 0.98 & 3.59 & 3.61 & 4.35 & 2.82 & 0.48 \\
\makecell{\puncttable} & 100k & 0.86 & 0.81 & 0.86 & 0.92 & 3.59 & 3.35 & 4.42 & 2.99 & 0.46 \\
\makecell{GPT-4 \cite{openai2023gpt4}} & 100k & 0.85 & 0.75 & 0.84 & 0.95 & 3.68 & 3.59 & 4.54 & 2.92 & 0.47 \\
\makecell{\gpt} & 100k & 0.85 & 0.74 & 0.86 & 0.94 & 3.67 & 3.66 & 4.43 & 2.92 & 0.48 \\
\makecell{\puncttable} & 128k & 0.85 & 0.80 & 0.85 & 0.89 & 3.66 & 3.38 & 4.49 & 3.10 & 0.45 \\
\makecell{\gpt} & 128k & 0.83 & 0.73 & 0.85 & 0.91 & 3.75 & 3.71 & 4.50 & 3.03 & 0.47 \\
\makecell{\identitytable} & 64k & 0.82 & 0.63 & 0.79 & 1.04 & 3.94 & 4.36 & 4.83 & 2.64 & 0.61 \\
\makecell{Bloom \cite{bloom}} & 250k & 0.81 & 0.77 & 0.84 & 0.81 & 3.91 & 3.49 & 4.56 & 3.67 & 0.46 \\
\makecell{\identitytable} & 80k & 0.80 & 0.61 & 0.76 & 1.01 & 4.07 & 4.50 & 5.00 & 2.71 & 0.61 \\
\makecell{\gpt } & 256k & 0.78 & 0.71 & 0.82 & 0.82 & 3.95 & 3.83 & 4.66 & 3.35 & 0.47 \\
\makecell{\identitytable} & 100k & 0.77 & 0.59 & 0.74 & 0.98 & 4.20 & 4.63 & 5.16 & 2.80 & 0.61 \\
\bottomrule
\end{tabular}
}
\caption{Compression metrics for tokenizers calculated on Code, English, and Multilingual datasets. Normalized Sequence Length is a ratio calculated against the llama tokenizer, and indicates the average sequence length that a tokenizer would have compared to llama. Bytes per tokens measures the number of UTF-8 bytes per (tokenized) token, where a higher Bytes per token would imply a greater level of compression.
Average Renyi \cite{Zouhar_Meister_Gastaldi_Du_Sachan_Cotterell_2023} is a distribution metric that measures how evenly the tokenizer allocates its vocabulary for a given text, it is calculated here with $\alpha = 2.5$ (see discussion in Appendix~\ref{sec:renyi}).
}
\label{tab:code-compression-all}
\end{table*}

\begin{table*}[h]
\centering
\scalebox{0.8}{
\begin{tabularx}{\linewidth}{|l|X|}
\hline
\multirow{7}{*}{Original snippet} & import numpy as np \\
& from tqdm import tqdm \\
& \\
& \# digits: 1000 \\
& l = [] \\
& for i in range(1000): \\
& \hspace*{4mm} l.append(str(i).zfill(3))) \\
\hline
\multirow{5}{*}{\shortstack{GPT-2 \\ \cite{gpt2} \\ Encoded length: 52}} &
\textcolor{blue}{\textbackslash n}\\ &\textcolor{red}{import}\textcolor{blue}{\textbullet n}\textcolor{red}{umpy}\textcolor{blue}{\textbullet as}\textcolor{red}{\textbullet np}\textcolor{blue}{\textbackslash n}\\ &\textcolor{red}{from}\textcolor{blue}{\textbullet t}\textcolor{red}{q}\textcolor{blue}{dm}\textcolor{red}{\textbullet import}\textcolor{blue}{\textbullet t}\textcolor{red}{q}\textcolor{blue}{dm}\textcolor{red}{\textbackslash n}\\ &\textcolor{blue}{\textbackslash n}\\ &\textcolor{red}{\#}\textcolor{blue}{\textbullet digits}\textcolor{red}{:}\textcolor{blue}{\textbullet 1000}\textcolor{red}{\textbullet }\textcolor{blue}{\textbackslash n}\\ &\textcolor{red}{l}\textcolor{blue}{\textbullet =}\textcolor{red}{\textbullet []}\textcolor{blue}{\textbackslash n}\\ &\textcolor{red}{for}\textcolor{blue}{\textbullet i}\textcolor{red}{\textbullet in}\textcolor{blue}{\textbullet range}\textcolor{red}{(}\textcolor{blue}{1000}\textcolor{red}{):}\textcolor{blue}{\textbackslash n}\\ &\textcolor{red}{\textbullet }\textcolor{blue}{\textbullet }\textcolor{red}{\textbullet }\textcolor{blue}{\textbullet l}\textcolor{red}{.}\textcolor{blue}{append}\textcolor{red}{(}\textcolor{blue}{str}\textcolor{red}{(}\textcolor{blue}{i}\textcolor{red}{).}\textcolor{blue}{z}\textcolor{red}{fill}\textcolor{blue}{(}\textcolor{red}{3}\textcolor{blue}{)))}\textcolor{red}{\textbackslash n}\\
\midrule
\multirow{5}{*}{\shortstack{\llama \\ \cite{touvron2023llama} \\ Encoded length: 57}} & \textcolor{blue}{\textbullet }\textcolor{red}{\textbackslash n}\\ &\textcolor{blue}{import}\textcolor{red}{\textbullet numpy}\textcolor{blue}{\textbullet as}\textcolor{red}{\textbullet np}\textcolor{blue}{\textbackslash n}\\ &\textcolor{red}{from}\textcolor{blue}{\textbullet t}\textcolor{red}{q}\textcolor{blue}{dm}\textcolor{red}{\textbullet import}\textcolor{blue}{\textbullet t}\textcolor{red}{q}\textcolor{blue}{dm}\textcolor{red}{\textbackslash n}\\ &\textcolor{blue}{\textbackslash n}\\ &\textcolor{red}{\#}\textcolor{blue}{\textbullet digits}\textcolor{red}{:}\textcolor{blue}{\textbullet }\textcolor{red}{1}\textcolor{blue}{0}\textcolor{red}{0}\textcolor{blue}{0}\textcolor{red}{\textbullet }\textcolor{blue}{\textbackslash n}\\ &\textcolor{red}{l}\textcolor{blue}{\textbullet =}\textcolor{red}{\textbullet []}\textcolor{blue}{\textbackslash n}\\ &\textcolor{red}{for}\textcolor{blue}{\textbullet i}\textcolor{red}{\textbullet in}\textcolor{blue}{\textbullet range}\textcolor{red}{(}\textcolor{blue}{1}\textcolor{red}{0}\textcolor{blue}{0}\textcolor{red}{0}\textcolor{blue}{):}\textcolor{red}{\textbackslash n}\\ &\textcolor{blue}{\textbullet \textbullet \textbullet }\textcolor{red}{\textbullet l}\textcolor{blue}{.}\textcolor{red}{append}\textcolor{blue}{(}\textcolor{red}{str}\textcolor{blue}{(}\textcolor{red}{i}\textcolor{blue}{).}\textcolor{red}{z}\textcolor{blue}{fill}\textcolor{red}{(}\textcolor{blue}{3}\textcolor{red}{)))}\textcolor{blue}{\textbackslash n}\\
\midrule
\multirow{5}{*}{\shortstack{\gpt \\ \cite{openai2023gpt4} \\ Encoded length: 40}} & 
\textcolor{blue}{\textbackslash n}\\ &\textcolor{red}{import}\textcolor{blue}{\textbullet numpy}\textcolor{red}{\textbullet as}\textcolor{blue}{\textbullet np}\textcolor{red}{\textbackslash n}\\ &\textcolor{blue}{from}\textcolor{red}{\textbullet tqdm}\textcolor{blue}{\textbullet import}\textcolor{red}{\textbullet tqdm}\textcolor{blue}{\textbackslash n\textbackslash n}\\ &\\ &\textcolor{red}{\#}\textcolor{blue}{\textbullet digits}\textcolor{red}{:}\textcolor{blue}{\textbullet }\textcolor{red}{100}\textcolor{blue}{0}\textcolor{red}{\textbullet \textbackslash n}\\ &\textcolor{blue}{l}\textcolor{red}{\textbullet =}\textcolor{blue}{\textbullet []\textbackslash n}\\ &\textcolor{red}{for}\textcolor{blue}{\textbullet i}\textcolor{red}{\textbullet in}\textcolor{blue}{\textbullet range}\textcolor{red}{(}\textcolor{blue}{100}\textcolor{red}{0}\textcolor{blue}{):\textbackslash n}\\ &\textcolor{red}{\textbullet \textbullet \textbullet }\textcolor{blue}{\textbullet l}\textcolor{red}{.append}\textcolor{blue}{(str}\textcolor{red}{(i}\textcolor{blue}{).}\textcolor{red}{z}\textcolor{blue}{fill}\textcolor{red}{(}\textcolor{blue}{3}\textcolor{red}{)))\textbackslash n}\\ 
\midrule
\multirow{5}{*}{\shortstack{\identitytable 100k \\ Encoded length: 25}} & \textcolor{blue}{\textbackslash nimport\textbullet numpy\textbullet as\textbullet np}\\ &\textcolor{red}{\textbackslash nfrom\textbullet }\textcolor{blue}{tq}\textcolor{red}{d}\textcolor{blue}{m\textbullet }\textcolor{red}{import\textbullet }\textcolor{blue}{tqdm} \\ & \\&\textcolor{red}{\textbackslash n\textbackslash n\#\textbullet }\textcolor{blue}{digit}\textcolor{red}{s:\textbullet }\textcolor{blue}{1000\textbullet }\textcolor{red}{\textbackslash n}\\ &\textcolor{blue}{l\textbullet =\textbullet }\textcolor{red}{[]\textbackslash n}\\ &\textcolor{blue}{for\textbullet i\textbullet in\textbullet range(}\textcolor{red}{1000}\textcolor{blue}{):\textbackslash n\textbullet \textbullet \textbullet \textbullet }\\&\textcolor{red}{l.}\textcolor{blue}{append(}\textcolor{red}{str}\textcolor{blue}{(i).}\textcolor{red}{z}\textcolor{blue}{fill}\textcolor{red}{(3}\textcolor{blue}{)))\textbackslash n}\\
\midrule
\multirow{5}{*}{\shortstack{\gpt 100k \\ Encoded length: 40}} & 
\textcolor{blue}{\textbackslash n}\\ &\textcolor{red}{import}\textcolor{blue}{\textbullet numpy}\textcolor{red}{\textbullet as}\textcolor{blue}{\textbullet np}\textcolor{red}{\textbackslash n}\\ &\textcolor{blue}{from}\textcolor{red}{\textbullet tqdm}\textcolor{blue}{\textbullet import}\textcolor{red}{\textbullet tqdm}\textcolor{blue}{\textbackslash n\textbackslash n}\\ &\\ &\textcolor{red}{\#}\textcolor{blue}{\textbullet digits}\textcolor{red}{:}\textcolor{blue}{\textbullet }\textcolor{red}{100}\textcolor{blue}{0}\textcolor{red}{\textbullet \textbackslash n}\\ &\textcolor{blue}{l}\textcolor{red}{\textbullet =}\textcolor{blue}{\textbullet []\textbackslash n}\\ &\textcolor{red}{for}\textcolor{blue}{\textbullet i}\textcolor{red}{\textbullet in}\textcolor{blue}{\textbullet range}\textcolor{red}{(}\textcolor{blue}{100}\textcolor{red}{0}\textcolor{blue}{):\textbackslash n}\\ &\textcolor{red}{\textbullet \textbullet \textbullet }\textcolor{blue}{\textbullet l}\textcolor{red}{.append}\textcolor{blue}{(str}\textcolor{red}{(i}\textcolor{blue}{).}\textcolor{red}{z}\textcolor{blue}{fill}\textcolor{red}{(}\textcolor{blue}{3}\textcolor{red}{)))\textbackslash n}\\
\midrule
\multirow{5}{*}{\shortstack{\puncttable 100k \\ Encoded length: 43}} & 
\textcolor{blue}{\textbackslash n}\\ &\textcolor{red}{import}\textcolor{blue}{\textbullet numpy}\textcolor{red}{\textbullet as}\textcolor{blue}{\textbullet np}\textcolor{red}{\textbackslash n}\\ &\textcolor{blue}{from}\textcolor{red}{\textbullet tqdm}\textcolor{blue}{\textbullet import}\textcolor{red}{\textbullet tqdm}\textcolor{blue}{\textbackslash n\textbackslash n}\\ &\\ &\textcolor{red}{\#}\textcolor{blue}{\textbullet digits}\textcolor{red}{:}\textcolor{blue}{\textbullet }\textcolor{red}{100}\textcolor{blue}{0}\textcolor{red}{\textbullet \textbackslash n}\\ &\textcolor{blue}{l}\textcolor{red}{\textbullet =}\textcolor{blue}{\textbullet []\textbackslash n}\\ &\textcolor{red}{for}\textcolor{blue}{\textbullet i}\textcolor{red}{\textbullet in}\textcolor{blue}{\textbullet range}\textcolor{red}{(}\textcolor{blue}{100}\textcolor{red}{0}\textcolor{blue}{):\textbackslash n}\\ &\textcolor{red}{\textbullet \textbullet \textbullet }\textcolor{blue}{\textbullet l}\textcolor{red}{.}\textcolor{blue}{append}\textcolor{red}{(}\textcolor{blue}{str}\textcolor{red}{(}\textcolor{blue}{i}\textcolor{red}{).}\textcolor{blue}{z}\textcolor{red}{fill}\textcolor{blue}{(}\textcolor{red}{3}\textcolor{blue}{)))\textbackslash n}\\
 \midrule
\multirow{5}{*}{\shortstack{\mergedtable 80k \\ Encoded length: 44}} & 
\textcolor{blue}{\textbullet \textbackslash n}\\ &\textcolor{red}{import}\textcolor{blue}{\textbullet numpy}\textcolor{red}{\textbullet as}\textcolor{blue}{\textbullet np}\textcolor{red}{\textbackslash n}\\ &\textcolor{blue}{from}\textcolor{red}{\textbullet tqdm}\textcolor{blue}{\textbullet import}\textcolor{red}{\textbullet tqdm}\textcolor{blue}{\textbackslash n\textbackslash n}\\ &\\ &\textcolor{red}{\#}\textcolor{blue}{\textbullet digits}\textcolor{red}{:}\textcolor{blue}{\textbullet }\textcolor{red}{1}\textcolor{blue}{0}\textcolor{red}{0}\textcolor{blue}{0}\textcolor{red}{\textbullet \textbackslash n}\\ &\textcolor{blue}{l}\textcolor{red}{\textbullet =}\textcolor{blue}{\textbullet []\textbackslash n}\\ &\textcolor{red}{for}\textcolor{blue}{\textbullet i}\textcolor{red}{\textbullet in}\textcolor{blue}{\textbullet range}\textcolor{red}{(}\textcolor{blue}{1}\textcolor{red}{0}\textcolor{blue}{0}\textcolor{red}{0}\textcolor{blue}{):\textbackslash n}\\ &\textcolor{red}{\textbullet \textbullet \textbullet \textbullet }\textcolor{blue}{l}\textcolor{red}{.append}\textcolor{blue}{(str}\textcolor{red}{(i}\textcolor{blue}{).}\textcolor{red}{z}\textcolor{blue}{fill}\textcolor{red}{(}\textcolor{blue}{3}\textcolor{red}{)))\textbackslash n}\\
\hline
\end{tabularx}}
\caption{Comparison of tokenizer encoding for a code sample. Spaces in the tokens are indicated with \textbullet, token delimitations are indicated by alternating colors.}
\label{tab:encodings}
\end{table*}

\begin{table*}[h]
\centering
\begin{tabular}{lr|r|r|r|r|r}
\toprule \\
 & \multicolumn{6}{c}{\textbf{Token-Equivalent (500B tokens)}} \\
 & \multicolumn{3}{c}{\textbf{HumanEval}} & \multicolumn{3}{c}{\textbf{MBPP}} \\
 & \multicolumn{1}{l}{\small{\textbf{Pass@1}}} & \multicolumn{1}{l}{\small{\textbf{Pass@100}}} & \multicolumn{1}{l}{\small{\textbf{Compile@1}}} & \multicolumn{1}{l}{\small{\textbf{Pass@1}}} & \multicolumn{1}{l}{\small{\textbf{Pass@100}}} & \multicolumn{1}{l}{\small{\textbf{Compile@1}}} \\
\toprule 
\small{\identitytable}\textsuperscript{R} & \acc{15.82621951}&\acc{49.90647628}& \acc{96.50609756}& \acc{21.72} & \acc{66.69038281} & \acc{98.967}\\
\small{\puncttable}\textsuperscript{R} & \acc{18.2} & \acc{58.7} & \acc{97.4} & \acc{24.6} & \acc{71.2} & \acc{99.5} \\
\small{\gpt}\textsuperscript{R} & \acc{17.80182927} & \acc{54.6479909} & \acc{98.967} & \acc{25.229} & \acc{70.71734375} & \acc{99.119}\\
\small{\identitytable}\textsuperscript{NL} & \acc{17.95731707} & \acc{60.0986209} & \acc{97.4695122} & \acc{24.544} & \acc{67.19198438} & \acc{99.055}\\
\small{\llama}\textsuperscript{NL} & \acc{20.5} & \acc{66.1} & \acc{98.4} & \acc{28.0} & \acc{71.2} & \acc{99.4} \\
\small{\puncttable}\textsuperscript{NL} & \acc{21.1} & \acc{63.5} & \acc{97.6} & \acc{28.6} & \acc{72.6} & \acc{99.5} \\
\small{\gpt}\textsuperscript{NL} & \acc{20.5} & \acc{65.3} & \acc{98.0} & \acc{27.2} & \acc{70.8} & \acc{99.4} \\
\small{\identitytable}\textsuperscript{C} & \acc{18.8} & \acc{66.1} & \acc{96.7} & \acc{23.4} & \acc{67.8} & \acc{99.4} \\
\small{\llama}\textsuperscript{C} & \acc{22.8} & \acc{68.6} & \acc{98.8} & \acc{30.1} & \acc{74.4} & \acc{99.4} \\
\small{\puncttable}\textsuperscript{C} & \acc{22.2} & \acc{68.7} & \acc{98.1} & \acc{29.8} & \acc{73.6} & \acc{99.4} \\
\small{\gpt}\textsuperscript{C} & \acc{21.2} & \acc{67.5} & \acc{97.7} & \acc{28.9} & \acc{74.9} & \acc{99.4} \\
\toprule \\
 & \multicolumn{6}{c}{\textbf{Word-Equivalent}} \\
 & \multicolumn{3}{c}{\textbf{HumanEval}} & \multicolumn{3}{c}{\textbf{MBPP}} \\
 & \multicolumn{1}{l}{\small{\textbf{Pass@1}}} & \multicolumn{1}{l}{\small{\textbf{Pass@100}}} & \multicolumn{1}{l}{\small{\textbf{Compile@1}}} & \multicolumn{1}{l}{\small{\textbf{Pass@1}}} & \multicolumn{1}{l}{\small{\textbf{Pass@100}}} & \multicolumn{1}{l}{\small{\textbf{Compile@1}}} \\
 \midrule
 \small{\identitytable}\textsuperscript{R} & \acc{14.87195122} & \acc{47.54663681} & \acc{96.30182927} & \acc{19.443} & \acc{64.91213281} & \acc{98.682} \\
 \small{\puncttable}\textsuperscript{R} & \acc{17.6} & \acc{55.5} & \acc{96.9} & \acc{24.4} & \acc{71.0} & \acc{99.4} \\
\small{\gpt}\textsuperscript{R} & \acc{17.13414634} & \acc{53.72724728} & \acc{97.18902439} & \acc{23.914} & \acc{69.95092188} & \acc{99.146} \\
\small{\identitytable}\textsuperscript{NL} & \acc{17.07621951} & \acc{56.11781869} & \acc{97.50609756} & \acc{22.686} & \acc{67.22416406} & \acc{98.237} \\
\small{\llama}\textsuperscript{NL} & \acc{20.5} & \acc{66.1} & \acc{98.4} & \acc{28.0} & \acc{71.2} & \acc{99.4} \\
\small{\puncttable}\textsuperscript{NL} & \acc{20.7} & \acc{62.1} & \acc{97.1} & \acc{28.2} & \acc{72.0} & \acc{99.4} \\
\small{\gpt}\textsuperscript{NL} & \acc{19.7} & \acc{63.7} & \acc{97.8} & \acc{26.8} & \acc{69.1} & \acc{99.4} \\
\small{\identitytable}\textsuperscript{C} & \acc{16.5} & \acc{61.8} & \acc{95.4} & \acc{22.3} & \acc{66.9} & \acc{99.2} \\
\small{\llama}\textsuperscript{C} & \acc{22.8} & \acc{68.6} & \acc{98.8} & \acc{30.1} & \acc{74.4} & \acc{99.4} \\
\small{\puncttable}\textsuperscript{C} & \acc{22.3} & \acc{67.3} & \acc{97.3} & \acc{28.9} & \acc{71.8} & \acc{99.4} \\
\small{\gpt}\textsuperscript{C} & \acc{21.4} & \acc{67.3} & \acc{97.8} & \acc{28.5} & \acc{74.2} & \acc{99.4} \\
\bottomrule
\end{tabular}

\caption{
We report the performance of fine-tuned 1.5B models using different tokenizers and base models on our two task generation tasks (HumanEval and MBPP) at both Token-equivalent levels to the baseline and Word-Equivalent. 
For the Token-equivalent performance, this table repeats from Table~\ref{tab:finetuning-32k}.
All tokenizers presented here are of vocabulary size 32k -- see Table~\ref{tab:NSL} for compression statistics for each. \textsuperscript{R} indicates a random initialization, \textsuperscript{NL} indicates that the base model used is our \smallmodel~1.5B model, and \textsuperscript{C} indicates that the base model is \smallcodemodel~1.5B.
}
\label{tab:finetuning-32k-all}
\end{table*}

\end{document}